\documentclass[10pt,twocolumn,letterpaper]{article}

\usepackage{cvpr}              

\makeatletter
\renewcommand{\paragraph}{%
  \@startsection{paragraph}{4}{\z@}%
  {0.4em}
  {-0.8em}
  {\normalfont\normalsize\bfseries}%
}
\makeatother

\usepackage{epsfig}
\usepackage{array}
\usepackage{tabularx}
\usepackage{makecell}
\usepackage{comment}
\usepackage{tikz}
\usetikzlibrary{arrows.meta,positioning,calc}
\usepackage{cuted}
\usepackage[accsupp]{axessibility}

\usepackage{microtype}
\usepackage{colortbl}
\usepackage{stfloats}

\usepackage{soul}
\usepackage[disable]{todonotes}

{\vspace{-\topsep}\begin{itemize}\itemsep1pt \parskip0pt \parsep0pt}
{\end{itemize}\vspace{-\topsep}}

\definecolor{gold(bg)}{rgb}{0.87, 0.79, 0.32}
\definecolor{gold(metallic)}{rgb}{0.83, 0.69, 0.22}
\definecolor{gold(web)(golden)}{rgb}{1.0, 0.84, 0.0}

\definecolor{palesilver}{rgb}{0.79, 0.75, 0.73}
\definecolor{silver}{rgb}{0.75, 0.75, 0.75}
\definecolor{lightslategray}{rgb}{0.47, 0.53, 0.6}

\definecolor{bronze(bg)}{rgb}{0.9, 0.6, 0.3}
\definecolor{bronze}{rgb}{0.8, 0.5, 0.2}

\definecolor{goldL}{HTML}{FBF2D2}
\definecolor{silverL}{HTML}{DDDDDD}
\definecolor{bronzeL}{HTML}{EED2B8}

\definecolor{goldD}{HTML}{D9AE13}
\definecolor{silverD}{HTML}{909090}
\definecolor{bronzeD}{HTML}{9A5F26}

\newcommand*\circledd[3]{\tikz[baseline=(char.base)]{
            \node[shape=circle,fill=#1,draw=#2,inner sep=1pt] (char) {\small{#3}};}}

\newcommand{\mfirst}[1]{%
    {#1\raisebox{0.8pt}{\circledd{goldL}{goldD}{1}}}%
}
\newcommand{\msecond}[1]{%
    {#1\raisebox{0.8pt}{\circledd{silverL}{silverD}{2}}}%
}
\newcommand{\mthird}[1]{%
    {#1\raisebox{0.8pt}{\circledd{bronzeL}{bronzeD}{3}}}%
}

\definecolor{mygray}{gray}{0.8}
\newcommand{\oldfirst}[1]{%
    {#1\raisebox{0.8pt}{\circledd{mygray}{mygray}{1}}}%
}
\newcommand{\oldsecond}[1]{%
    {#1\raisebox{0.8pt}{\circledd{mygray}{mygray}{2}}}%
}
\newcommand{\oldthird}[1]{%
    {#1\raisebox{0.8pt}{\circledd{mygray}{mygray}{3}}}%
}

\newcommand{\medal}[3]{\tikz[baseline=(char.base)]{\node[rounded corners=2pt,fill=#1,draw=#2,inner sep=1.5pt] (char) {#3};}}

\newcommand{\bm}[2]{
    \ifcase#1\or
      {\medal{goldL}{goldD}{\textbf{#2}}}
    \or 
      {\medal{silverL}{silverD}{#2}}
    \or 
      {\medal{bronzeL}{bronzeD}{#2}}
    \else 
      #2
    \fi\ignorespaces
}

\newcolumntype{L}[1]{>{\raggedright\let\newline\\\arraybackslash\hspace{0pt}}b{#1}}
\newcolumntype{C}[1]{>{\centering\let\newline\\\arraybackslash\hspace{0pt}}b{#1}}

\newcommand{\rankn}[1]{({\small\##1})}

\def\paperID{*****} 
\def\confName{CVPR}
\def\confYear{2026}

\definecolor{cvprblue}{rgb}{0.21,0.49,0.74}
\usepackage[pagebackref,breaklinks,colorlinks,allcolors=cvprblue]{hyperref}

\begin{document}

\title{4\textsuperscript{th} Workshop on Maritime Computer Vision (MaCVi): Challenge Overview}

\author{
Benjamin Kiefer$^{1}$, Jan Lukas Augustin$^{2,3}$, Jon Muhovič$^{4}$, Mingi Jeong$^{5}$, Arnold Wiliem$^{6,7}$,\\
Janez Pers$^{4}$, Matej Kristan$^{4}$, Alberto Quattrini Li$^{8}$, Matija Teršek$^{9}$, Josip Šarić$^{10}$, Arpita Vats$^{11}$,\\
Dominik Hildebrand$^{12}$, Rafia Rahim$^{12}$, Mahmut Karaaslan$^{13}$, Arpit Vaishya$^{1}$, Steve Xie$^{1}$, Ersin Kaya$^{13}$, \\ Akib Mashrur$^{6}$,
Tze-Hsiang Tang$^{14}$, Chun-Ming Tsai$^{15}$, Jun-Wei Hsieh$^{16}$, Ming-Ching Chang$^{17,18}$,\\ Wonwoo Jo$^{19}$,
Doyeon Lee$^{20}$, Yusi Cao$^{21}$, Lingling Li$^{21}$, Vinayak Nageli$^{22}$, Arshad Jamal$^{23}$,\\
Gorthi Rama Krishna Sai Subrahmanyam$^{22}$, Jemo Maeng$^{24}$, Seongju Lee$^{24}$, Kyoobin Lee$^{24}$, Xu Liu$^{21}$,\\
LiCheng Jiao$^{21}$, Jannik Sheikh$^{25}$, Martin Weinmann$^{26}$, Ivan Martinović$^{10}$,\\
Jose Mateus Raitz Persch$^{27}$, Rahul Harsha Cheppally$^{27}$, Mehmet E. Belviranli$^{28}$, Dimitris Gahtidis$^{6}$,\\
Hyewon Chun$^{19}$, Sangmun Lee$^{19}$, Philipp Gorczak$^{3}$, Hansol Kim$^{19}$, Jeeyeon Jeon$^{19}$,\\
Borja Carrillo Perez$^{29}$, Jiahui Wang$^{21}$, Sangmin Park$^{24}$, Andreas Michel$^{25}$, Jannick Kuester$^{25}$,\\
Bettina Felten$^{25}$, Wolfgang Gross$^{25}$, Yuan Feng$^{30}$, Justin Davis$^{28}$\\
$^{1}$LOOKOUT, $^{2}$Helmut Schmidt University, $^{3}$catskill GmbH, $^{4}$University of Ljubljana, $^{5}$Virginia Tech,\\
$^{6}$Shield AI, $^{7}$Queensland University of Technology, $^{8}$Dartmouth College, $^{9}$Luxonis,\\
$^{10}$Faculty of Electrical Engineering and Computing, University of Zagreb, $^{11}$LinkedIn,\\
$^{12}$University of Tuebingen, $^{13}$Konya Technical University, $^{14}$Schneider Electric Taiwan Co., Ltd.,\\
$^{15}$University of Taipei, $^{16}$National Yang Ming Chiao Tung University, $^{17}$University at Albany, SUNY,\\
$^{18}$Inventec Corporation, $^{19}$HD Korea Shipbuilding \& Offshore Engineering Co., Ltd.,\\
$^{20}$Seoul National University, $^{21}$Xidian University, $^{22}$Indian Institute of Technology, Tirupati,\\
$^{23}$Centre for Artificial Intelligence and Robotics (CAIR), Bangalore, India,\\
$^{24}$Gwangju Institute of Science and Technology, $^{25}$Fraunhofer IOSB,\\
$^{26}$Karlsruhe Institute of Technology, $^{27}$Kansas State University, $^{28}$Colorado School of Mines,\\
$^{29}$Arquimea Research Center, $^{30}$Independent Researcher
}

\maketitle
\thispagestyle{empty}

\begin{abstract}
The 4$^{\text{th}}$ Workshop on Maritime Computer Vision (MaCVi) is organized as part of CVPR 2026. This edition features five benchmark challenges with emphasis on both predictive accuracy and embedded real-time feasibility. This report summarizes the MaCVi 2026 challenge setup, evaluation protocols, datasets, and benchmark tracks, and presents quantitative results, qualitative comparisons, and cross-challenge analyses of emerging method trends. We also include technical reports from top-performing teams to highlight practical design choices and lessons learned across the benchmark suite. Datasets, leaderboards, and challenge resources are available at

\url{https://macvi.org/workshop/cvpr26}.
\end{abstract}

\begin{strip}
\centering
\captionsetup{type=figure}
\captionsetup[subfigure]{font=footnotesize, justification=centering}
\begin{subfigure}[t]{.19\textwidth}
   \centering
   \includegraphics[height=1.7cm,trim=50 0 0 0,clip]{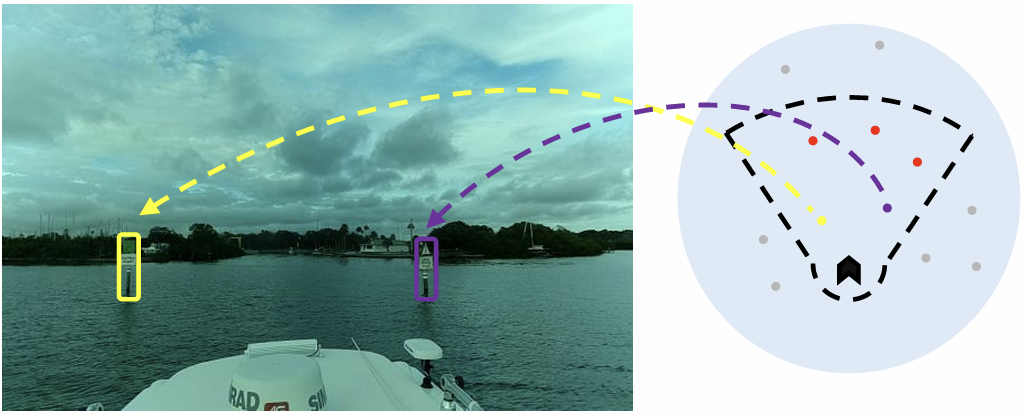}
   \caption{Vision-to-Chart}
   \label{fig:dist_estimation}
\end{subfigure}\hfill%
\begin{subfigure}[t]{.19\textwidth}
   \centering
   \includegraphics[height=1.7cm]{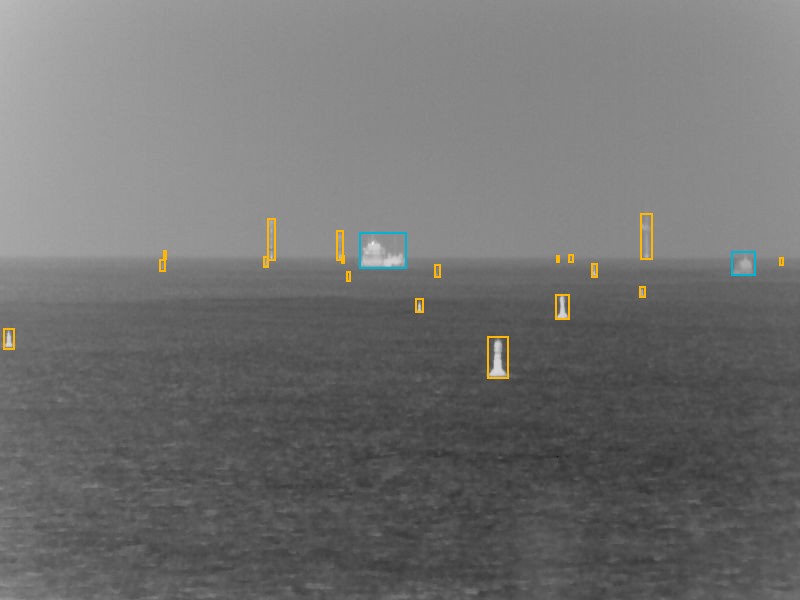}
   \caption{Thermal Detection}
   \label{fig:thermal_object_detection}
\end{subfigure}\hfill%
\begin{subfigure}[t]{.19\textwidth}
   \centering
   \includegraphics[height=1.7cm]{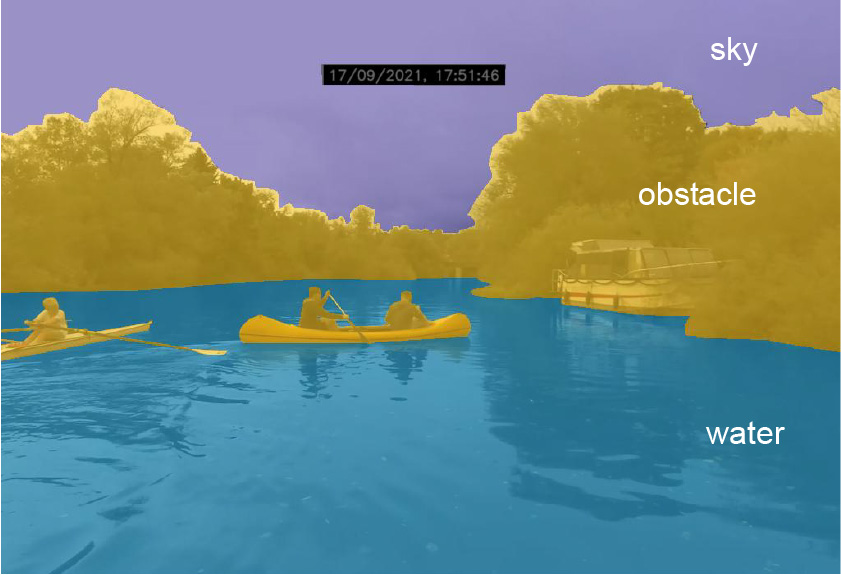}
   \caption{Embedded Seg.}
   \label{fig:obstacle_segmentation}
\end{subfigure}\hfill%
\begin{subfigure}[t]{.19\textwidth}
   \centering
   \includegraphics[height=1.7cm]{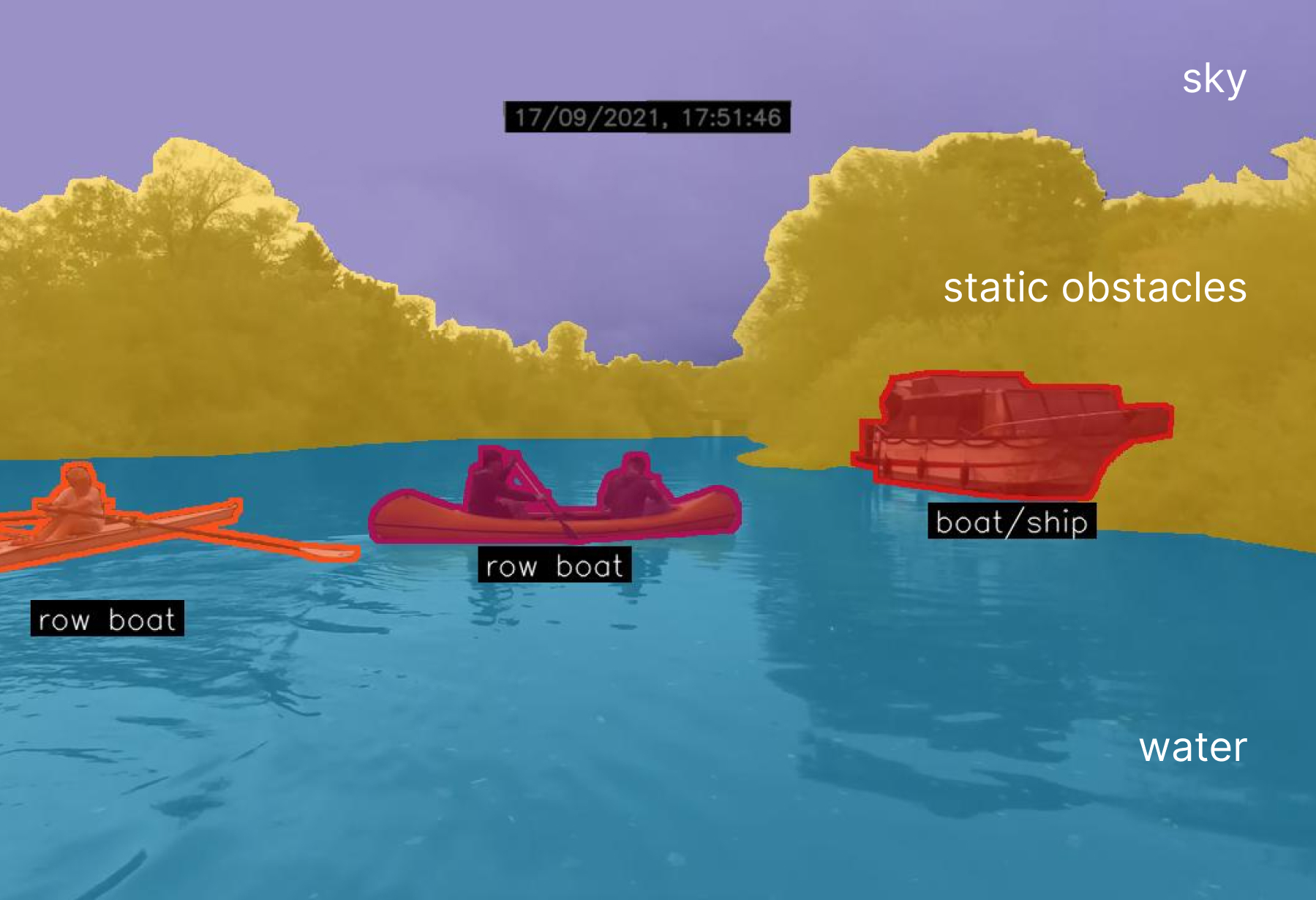}
   \caption{LaRS Panoptic}
   \label{fig:panoptic_segmentation}
\end{subfigure}\hfill%
\begin{subfigure}[t]{.19\textwidth}
   \centering
   \includegraphics[height=1.7cm]{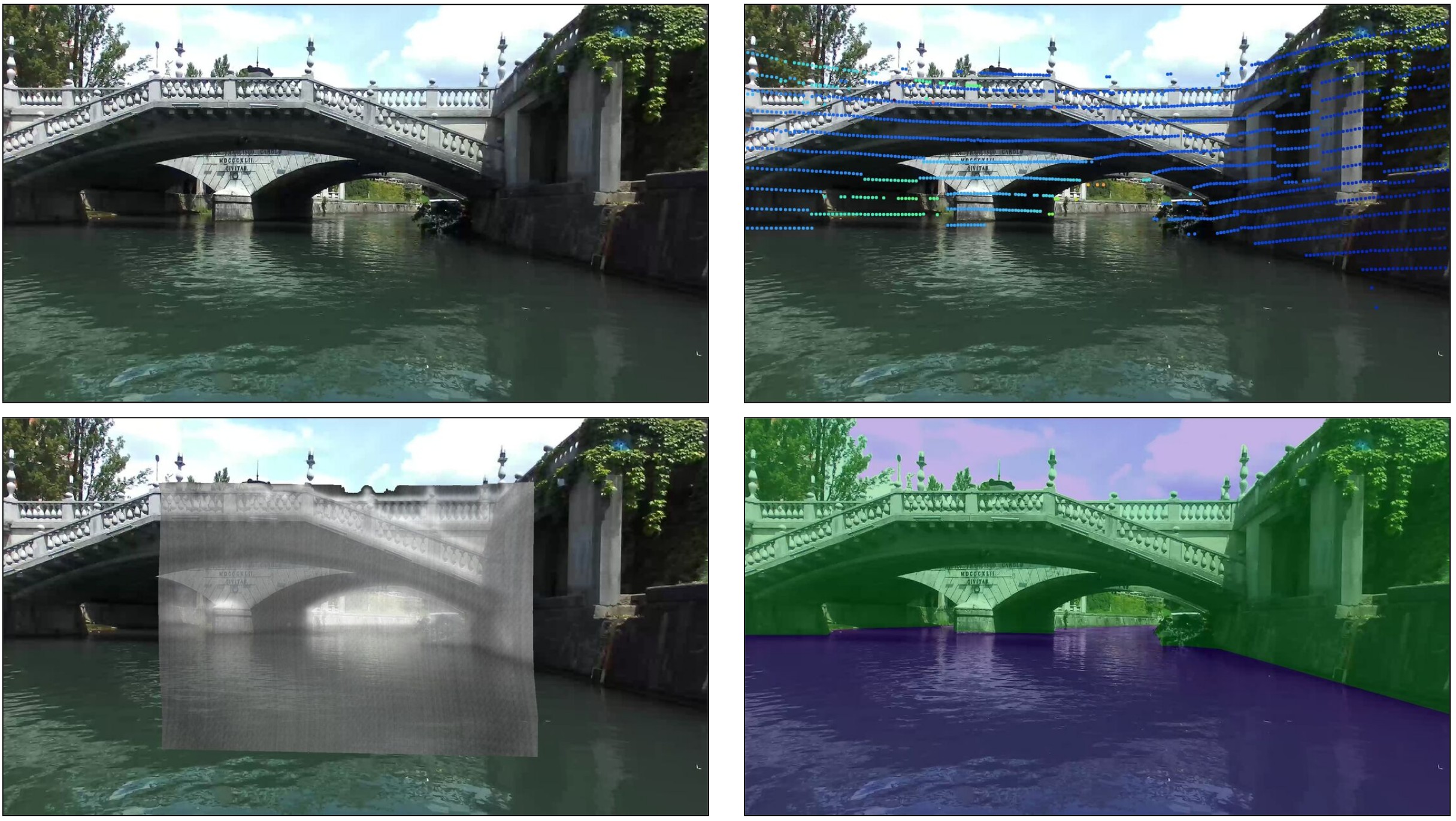}
   \caption{Multi-modal}
   \label{fig:generalist_meta}
\end{subfigure}
\addtocounter{figure}{-1}
\captionof{figure}{Overview of MaCVi @ CVPR 2026 challenges, including Vision-to-Chart Data Association, Thermal Object Detection Maritime Collision Avoidance Dataset, Embedded Semantic Segmentation, LaRS Panoptic Segmentation, and the Multi-modal Challenge Multiaqua.}
\label{fig:challenges_overview}
\end{strip}

\begin{figure*}[t]
    \centering
    \begin{subfigure}{0.245\linewidth}
        \centering
        \textbf{1st Place}\\[2pt]
        \includegraphics[width=\linewidth]{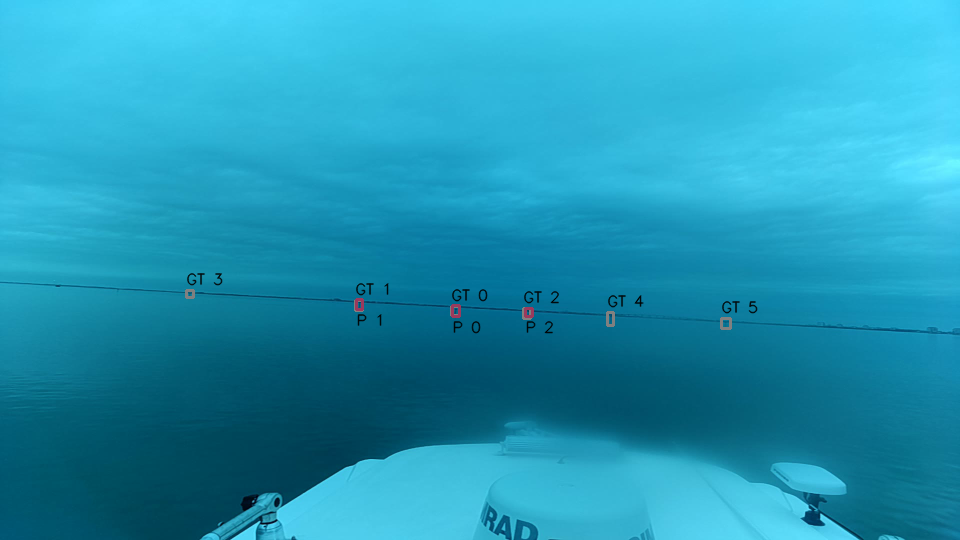}
    \end{subfigure}
    \hfill
    \begin{subfigure}{0.245\linewidth}
        \centering
        \textbf{2nd Place}\\[2pt]
        \includegraphics[width=\linewidth]{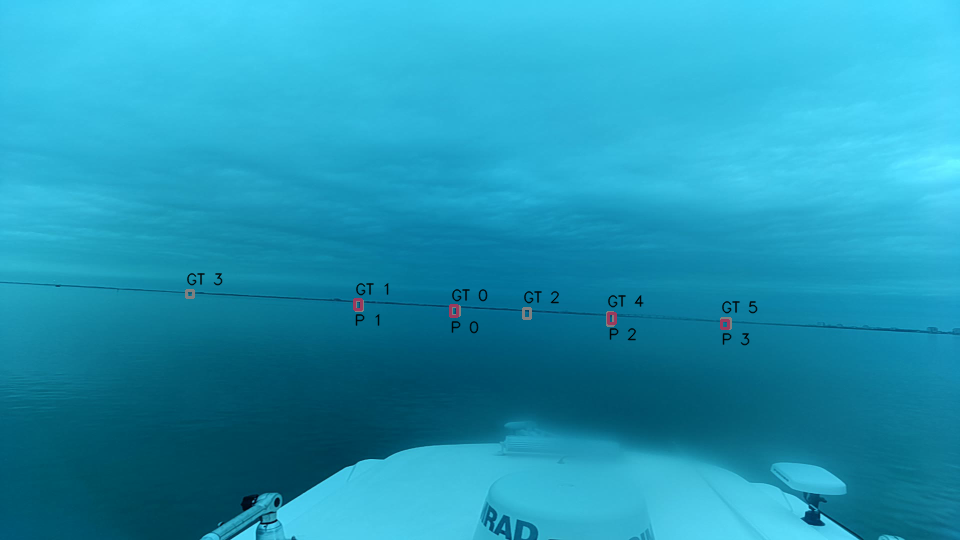}
    \end{subfigure}
    \hfill
    \begin{subfigure}{0.245\linewidth}
        \centering
        \textbf{3rd Place}\\[2pt]
        \includegraphics[width=\linewidth]{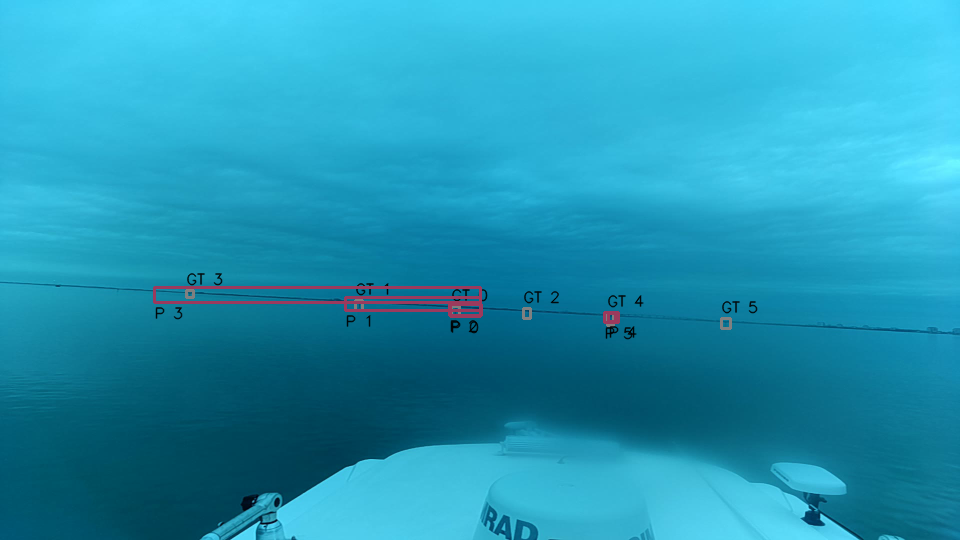}
    \end{subfigure}
    \hfill
    \begin{subfigure}{0.245\linewidth}
        \centering
        \textbf{4th Place}\\[2pt]
        \includegraphics[width=\linewidth]{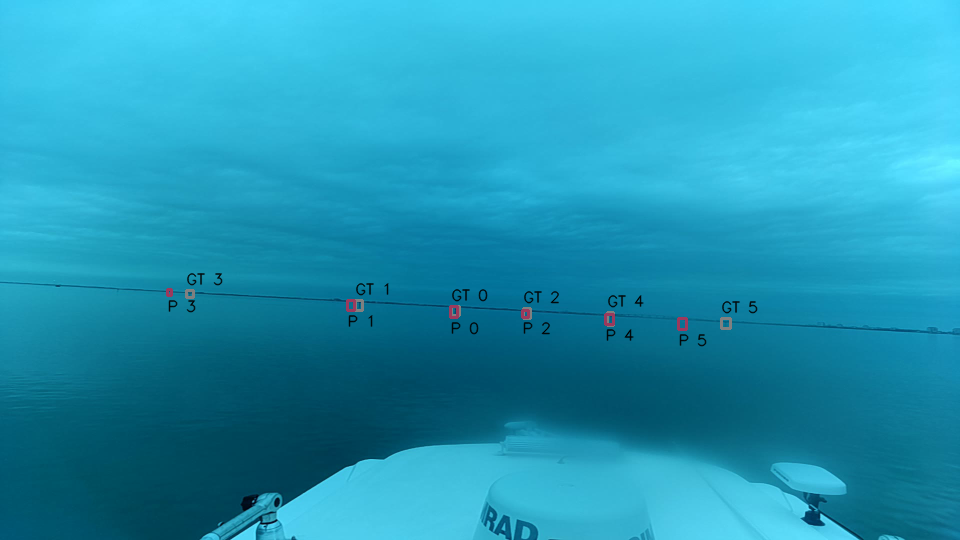}
    \end{subfigure}
    \caption{\textbf{Qualitative detection examples} from the Vision-to-Chart Data Association challenge. Each panel shows the overlay of predicted and ground-truth bounding boxes and predicted index of associated chart marker.}
    \label{fig:v2c-qualitative}
\end{figure*}

\begin{table*}[t]
\centering
\caption{Results for the Vision-to-Chart Data Association challenge on the held out test split. Ranking medals indicate final placement. The baseline (gray) is the organizer-provided DETR-style fusion transformer. Method names are derived from the submitted technical reports.}
\label{tab:v2c-results}
\resizebox{\linewidth}{!}{%
\begin{tabular}{cllccccc}
\toprule
Place & Method & Institution & P $\uparrow$ & R $\uparrow$ & F1 $\uparrow$ & mIoU $\uparrow$ & Overall $\uparrow$ \\
\midrule
\mfirst{} & Skyline-aware ROI-calibrated association [\S\ref{visionmap:skyline-roi}] & HD Korea Shipbuilding & \bm1{0.9248} & \bm2{0.6934} & \bm2{0.7926} & \bm1{0.7365} & \bm1{0.7646} \\
\msecond{} & QueryMLP projection + DETR [\S\ref{visionmap:querymlp}] & Arquimea Research Center & \bm2{0.8563} & \bm1{0.7604} & \bm1{0.8055} & \bm2{0.6718} & \bm2{0.7386} \\
\mthird{} & Dynamic chart-derived DEIMv2 [\S\ref{visionmap:deimv2-dynamic}] & Xidian University team & \bm3{0.4107} & \bm3{0.3827} & \bm3{0.3962} & \bm3{0.3705} & \bm3{0.3834} \\
- & \color{gray} Fusion transformer (baseline) & \color{gray} MaCVi Organizers & \color{gray} 0.3142 & \color{gray} 0.2925 & \color{gray} 0.3029 & \color{gray} 0.3636 & \color{gray} 0.3333 \\
4th & IMU-conditioned query DETR [\S\ref{visionmap:imu-detr}] & IIT Tirupati / CAIR-DRDO & 0.2170 & 0.2107 & 0.2138 & 0.2499 & 0.2318 \\

\bottomrule
\end{tabular}%
}
\end{table*}

\section{Introduction}
Maritime environments, with their unique challenges such as dynamic lighting, reflections, and cluttered scenes, demand specialized computer vision techniques \cite{Kiefer_2026_CVPR,varga2022seadronessee,prasad2019object,kanjir2018vessel,gallego2019detection,kiefer2021leveraging}. Autonomous systems like Unmanned Surface Vehicles (USVs) rely heavily on robust vision algorithms to navigate, detect, and interpret complex surroundings \cite{Zust2022Learning,Bovcon2021,kiefer2025approximatesupervisedobjectdistance}. Addressing these challenges requires not only cutting-edge algorithms but also standardized benchmarks and a collaborative research ecosystem \cite{KristanPAMI2016,bovcon2019mastr,rsuigm,bloisi2014background}.

The 4\textsuperscript{th} Workshop on Maritime Computer Vision (MaCVi 2026) builds on the momentum of previous iterations \cite{Kiefer_2023_WACV,Kiefer_2024_WACV,Kiefer_2025_WACV} and marks this year's workshop at CVPR 2026. The 2026 challenge suite includes Vision-to-Chart Data Association, Thermal Object Detection, LaRS Panoptic Segmentation, Embedded Semantic Segmentation, and Multimodal Semantic Segmentation.

This draft report summarizes the current challenge configuration for MaCVi 2026 and serves as the starting point for iterative updates. The rest of the paper is structured as follows: Section~\ref{sec:protocol} details the general challenge protocols, while Section~\ref{sec:challenge-tracks} outlines the challenge tracks. All datasets, evaluation tools, and leaderboards are available at \url{https://macvi.org/workshop/cvpr26}.

\section{Challenge Participation Protocol}
\label{sec:protocol}

As in previous challenge iterations, all tracks followed a shared participation and evaluation protocol, with task-specific details provided on the individual challenge pages. Rules, datasets, evaluation code, and starter kits were distributed through the workshop website, and participants submitted predictions through the official evaluation server in the required task-specific formats, including ONNX export where applicable. Submission frequency was limited to one to three entries per day depending on the track. The evaluation server handled automated scoring and public leaderboards for most tracks, with the challenge timeline spanning December~29,~2025 to March~15,~2026 (AoE). After final verification, teams were ranked by the official track metrics and compliance requirements, and top teams were invited to contribute short technical reports for this overview paper.

\section{MaCVi 2026 Challenge Tracks}
\label{sec:challenge-tracks}

MaCVi @ CVPR 2026 features five benchmark challenges. The 2026 edition puts explicit emphasis on both model quality and embedded real-time constraints.

\subsection{Vision-to-Chart Data Association Challenge}
\label{sec:vision-to-chart}

This challenge studies association between visible maritime navigational aids and chart markers from a monocular RGB image. Given an image and a set of chart queries, methods must detect visible buoys and assign them to the correct markers under clutter, reflections, and long-range viewpoints. The dataset contains 4,285 training, 904 validation, and 924 test samples; the training and validation splits are public, while the test split is private.

Ranking follows \cite{kreis2025realtimefusionvisualchart} and uses buoy-detection Precision/Recall/F1, matched-box mIoU, and an overall score defined as the arithmetic mean of F1 and mIoU. We use the real-time organizer baseline from \cite{kreis2025realtimefusionvisualchart}. Submissions were limited to 250M parameters and were evaluated by the organizers for reproducibility and private-test performance.

Table~\ref{tab:v2c-results} and Figure \ref{fig:v2c-qualitative} summarize the final results. All top-ranked methods outperform the organizer baseline, with the first two methods showing a substantial margin. Technical descriptions of all submissions are provided in Section~\ref{visionmap:submissions}.

Across submissions, the strongest trend is the use of explicit geometric priors. Top methods project chart information into image space, define query-specific search regions, or inject chart- and IMU-derived cues into the detector. The winning method combines skyline estimation, ROI projection, assignment, and calibration in a staged pipeline; the second-place method uses a learned world-to-image projection as a compact spatial prior; and the third-place method shows that a stronger visual backbone and decoder can further improve robustness. Overall, the results suggest that physically grounded geometric conditioning is especially effective for vision-to-chart association, and that stronger detection backbones remain complementary to these priors.

\begin{table*}[t]
\centering
\caption{Results for the Thermal Object Detection challenge. Ranking medals indicate final placement. The baseline (gray) is the organizer-provided Faster R-CNN with ResNet-50. Self-reported inference speed and hardware are included. Backbone and detection-level pretraining datasets are reported separately (IN = ImageNet-1K, IN-22K = ImageNet-22K, LVD = Large Vision Dataset).}
\label{tab:thermal-det-results}
\resizebox{\linewidth}{!}{%
\begin{tabular}{clllccccclll}
\toprule
Place & Method & Institution & AP $\uparrow$ & AP$_{50}$ & AP$_{75}$ & AR$_1$ & AR$_{10}$ & FPS & Hardware & Backbone pretr. & Det.\ pretr. \\
\midrule
\mfirst{} & Multi-arch ensemble + SSL [\S\ref{thermal-det:multi-arch-ssl}] & Schneider Electric Taiwan & \bm1{0.4868} & \bm1{0.8268} & \bm1{0.4709} & \bm1{0.3043} & \bm1{0.5937} & $\sim$0.01 & A100 40GB & IN-22K, LVD-142M & COCO, O365 \\
\msecond{} & DEIMv2 ensemble [\S\ref{thermal-det:deimv2-ensemble}] & U.\ Taipei / NYCU / UAlbany & \bm2{0.4709} & \bm2{0.8218} & \bm3{0.4409} & \bm3{0.2987} & \bm3{0.5763} & $\sim$0.1 & RTX 3090 & LVD-1689M & COCO \\
\mthird{} & AGAF [\S\ref{thermal-det:agaf}] & HD Korea Shipbuilding & \bm3{0.4685} & 0.8067 & \bm2{0.4628} & \bm2{0.3033} & \bm2{0.5850} & $\sim$1.5 & RTX 5070 & LVD-1689M & COCO \\
\midrule
- & \color{gray} Faster R-CNN (baseline) & \color{gray} MaCVi Organizers & \color{gray} 0.3137 & \color{gray} 0.6313 & \color{gray} 0.2769 & \color{gray} 0.2417 & \color{gray} 0.4167 & $\sim$32 & A30 & IN & COCO \\
\bottomrule
\end{tabular}}
\end{table*}

\begin{figure}[t]
    \centering
    \includegraphics[width=\linewidth]{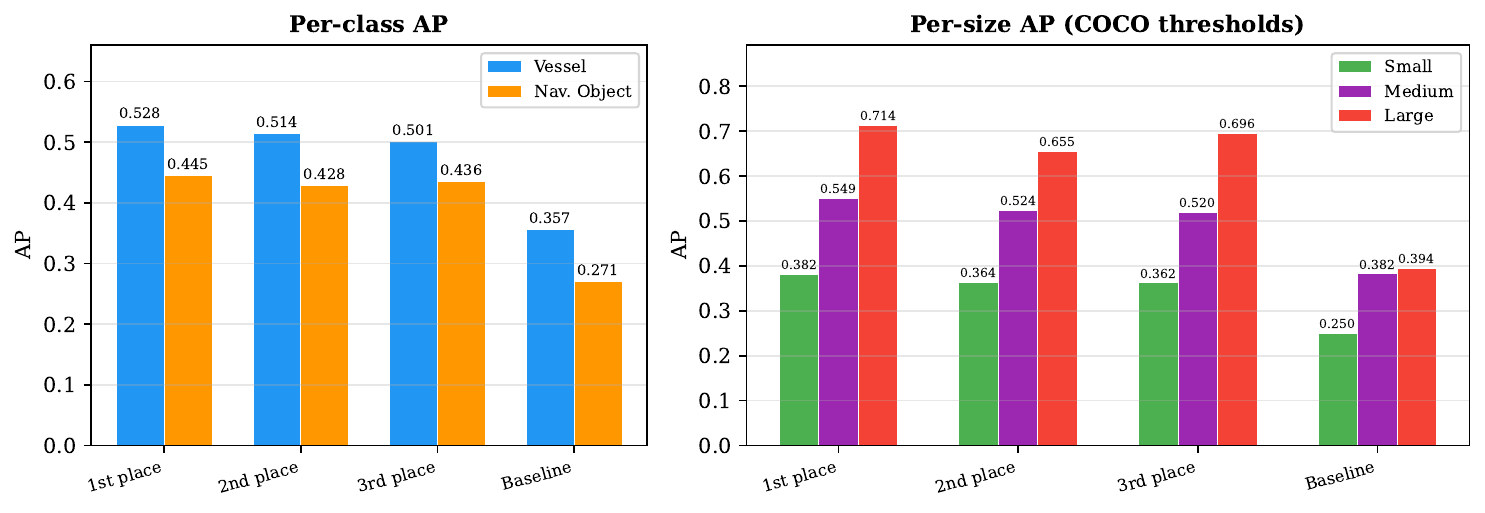}
    \caption{\textbf{Per-class and per-size AP breakdown} for the Thermal Object Detection challenge. Left: AP for \emph{vessel} vs.\ \emph{navigational object} classes. Right: AP at COCO size thresholds (small / medium / large). Vessel detection is consistently easier than navigational object detection across all methods. Small object AP remains the main bottleneck.}
    \label{fig:thermal-per-class}
\end{figure}

\subsection{Thermal Object Detection Challenge}
\label{sec:thermal-object-detection}

Night-time and low-visibility conditions are critical operational scenarios for unmanned surface vehicles, yet conventional RGB-based perception systems perform poorly in such conditions. Thermal infrared imaging offers a modality that is largely invariant to ambient illumination, making it a natural complement for maritime operations. This challenge targets obstacle and vessel detection in thermal imagery.

The challenge uses the Maritime Collision Avoidance Dataset~\cite{gorczak2025maritime}, which contains annotated electro-optical imagery captured from German, British, and Dutch waters between 2023 and 2025. For the MaCVi 2026 edition, the dataset is organized into two object classes (\emph{vessel} and \emph{navigational object}) with COCO-format bounding box annotations. The dataset comprises 704 train, 173 val, and 381 test images with hidden labels. An NVIDIA RTX 5080 GPU, sponsored by catskill GmbH, was offered as a prize for the top-performing team.

\subsubsection{Evaluation Protocol}

Detection performance is evaluated using the standard COCO object detection metrics. The primary ranking metric is Average Precision (AP), computed as AP averaged over IoU thresholds from 0.50 to 0.95 in steps of 0.05. AP$_{50}$ serves as a tiebreaker. Additional reported metrics include AP$_{75}$, AR$_1$, and AR$_{10}$.

Participants submit COCO-style JSON files containing bounding boxes, class labels, and confidence scores. A maximum of three submissions per day is allowed during the challenge period. All participants are required to report inference speed (in FPS) and the hardware used for benchmarking. As baseline, we provide a Faster R-CNN with a ResNet-50 backbone pretrained on COCO.

\begin{figure*}[t]
    \centering
    \includegraphics[width=\linewidth,trim={65pt 0 0 0},clip]{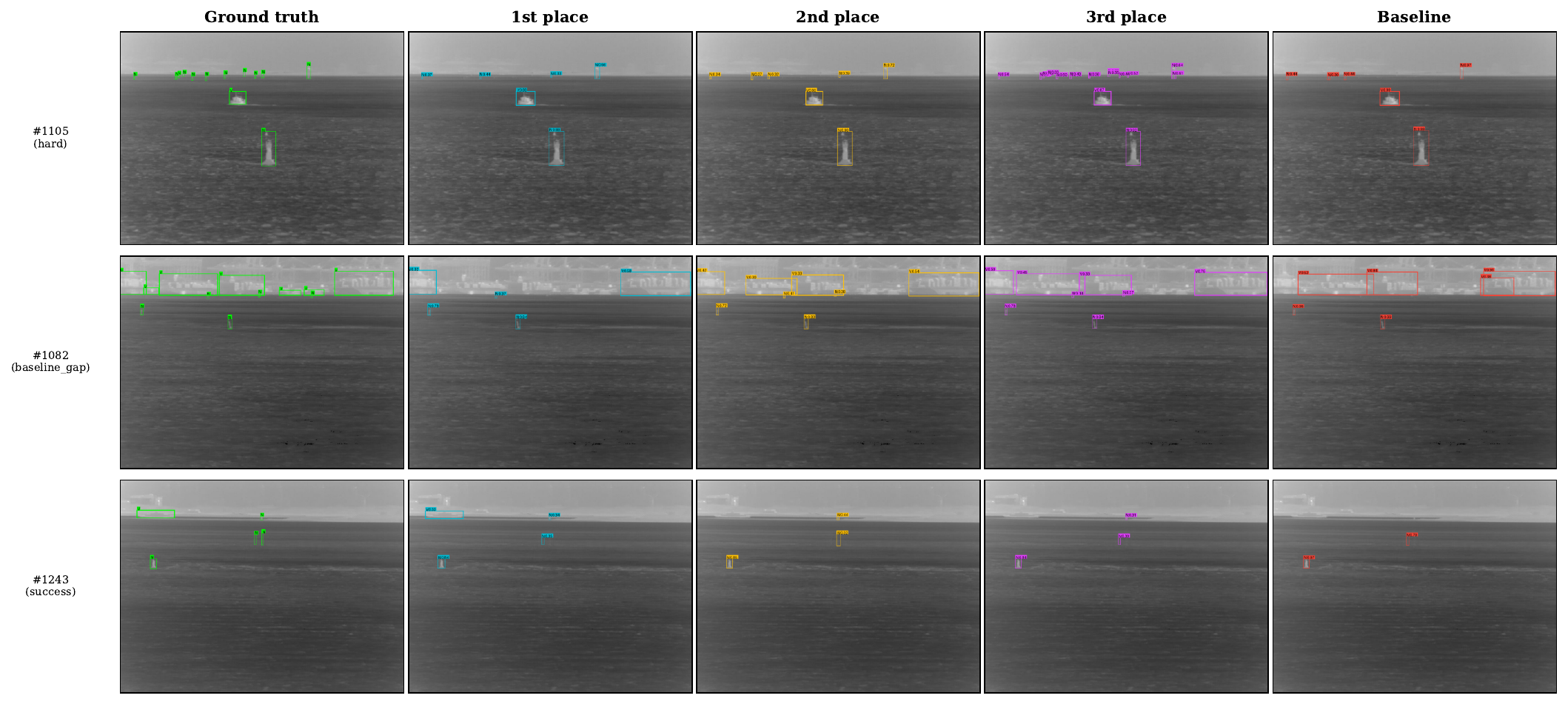}
    \caption{\textbf{Qualitative detection examples} from the Thermal Object Detection challenge. Each row shows one test image with ground truth (green) and predictions from the top three methods and baseline. Columns left to right: ground truth, 1st place, 2nd place, 3rd place, baseline. Predictions are filtered at confidence $\geq 0.3$. V = vessel, N = navigational object.}
    \label{fig:thermal-qualitative}
\end{figure*}

\subsubsection{Submissions, Analysis and Trends}

\begin{figure}[t]
    \centering
    \includegraphics[width=\linewidth]{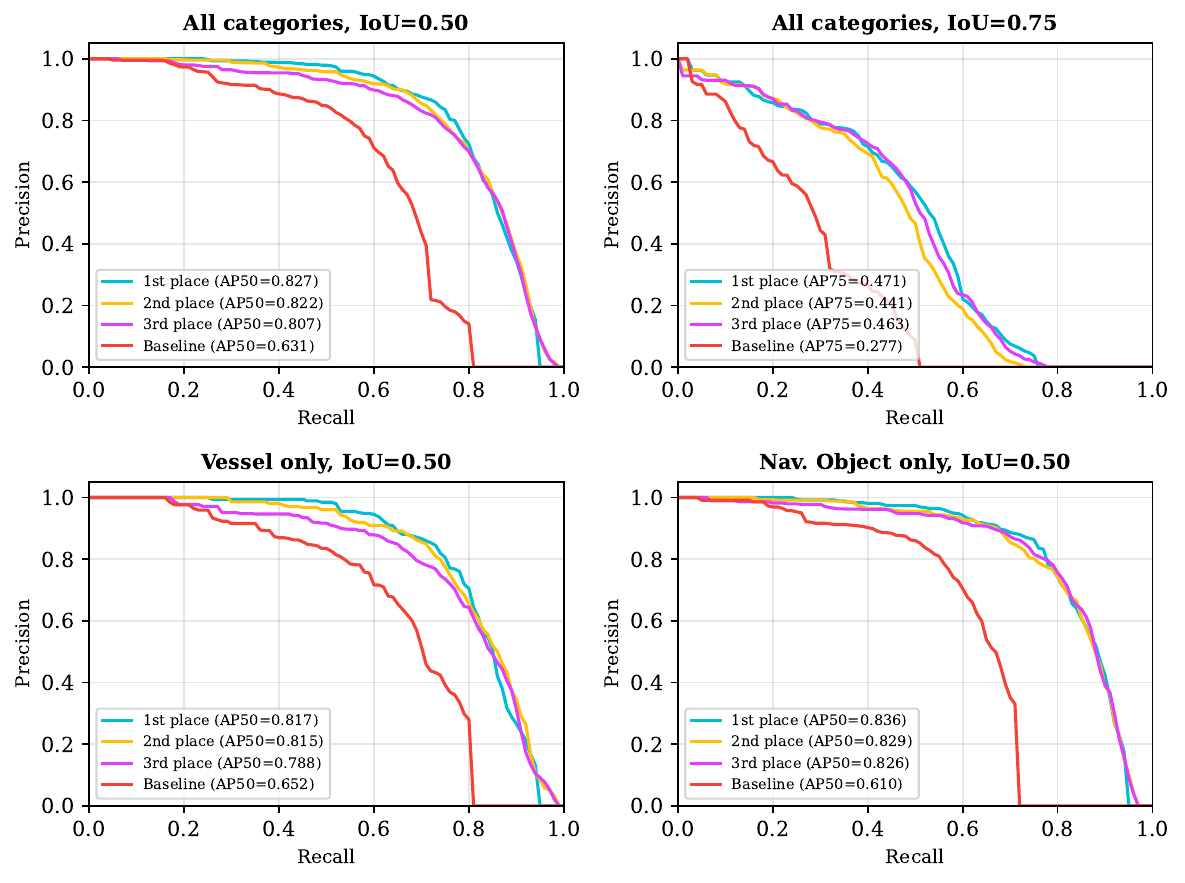}
    \caption{\textbf{Precision--Recall curves} for the Thermal Object Detection challenge. Curves are shown for each method at IoU=0.50.}
    \label{fig:thermal-pr-curves}
\end{figure}

Table~\ref{tab:thermal-det-results} summarizes the overall ranking, Figure~\ref{fig:thermal-per-class} reports the class and size breakdown, Figure~\ref{fig:thermal-qualitative} shows representative detections, and Figure~\ref{fig:thermal-pr-curves} plots the precision--recall curves at IoU~=~0.50.

\paragraph{Ensemble dominance.}
All three podium methods rely on multi-model ensembles fused via Weighted Boxes Fusion (WBF)~\cite{solovyev2021wbf}. The winning method combines 6 models spanning 5 distinct architectures (Co-DINO~\cite{zong2023codino}, DDQDETR~\cite{zhang2023ddqdetr}, DINO~\cite{zhang2023dino}, RTMDet~\cite{lyu2022rtmdet}, and RF-DETR~\cite{robinson2026rfdetr}), the 2nd place method ensembles 11 DEIMv2 detectors trained at 4 different resolutions with different random seeds, and the 3rd place method fuses 9 RF-DETR~2XLarge sources via class-aware WBF with per-class weights for vessels and navigational objects. The complementary strategies architecture diversity, optimization diversity, and class-aware fusion all prove effective for improving recall and localization stability.
\paragraph{Semi-supervised learning.}
The 1st place method employs a three-phase training pipeline that includes pseudo-labeling on the 381 unlabeled test images followed by MixPL~\cite{chen2023mixpl} teacher-student training. This semi-supervised component demonstrates that even modest amounts of unlabeled data can meaningfully improve performance when labeled training sets are small (704 images).

\paragraph{Resolution and small objects.}
A domain-specific challenge is that 63.4\% of annotated objects are smaller than $32 \times 32$ pixels (COCO ``small'' category), with navigational objects having a median size of only $8.3 \times 18.6$ pixels. The technical reports indicate that higher inference resolution is the single largest lever for improving detection of these small objects. The 2nd place method explores resolutions from $1024 \times 1024$ up to $1600 \times 1600$, assigning higher WBF weights to models operating at larger scales.

\paragraph{Contrast enhancement.}
Both top methods apply CLAHE (Contrast Limited Adaptive Histogram Equalization) to improve feature contrast in thermal imagery. Notably, the 1st place report observes that CLAHE benefits strong models but slightly hurts weaker ones ($-0.002$ to $-0.008$ AP), suggesting that it amplifies existing model capacity rather than providing a universal boost.

\paragraph{Annotation quality and domain-specific filtering.}
The 3rd place method demonstrates that carefully curating the training labels can rival more complex learning strategies. Fixing inconsistent wind-turbine annotations in the provided dataset yielded the single largest gain of any individual technique across all submissions (+0.027~AP). Additionally, a YOLO-based~\cite{Jocher_Ultralytics_YOLO_2023} horizon detector that suppresses navigational-object false positives above the estimated skyline contributed a further +0.012~AP. These domain-specific adaptations---annotation refinement and geometric filtering---are complementary to the ensemble and semi-supervised strategies employed by the other top teams.

\subsubsection{Discussion and Challenge Winners}

The winners of the Thermal Object Detection challenge are listed in Table \ref{tab:thermal-det-results}. All three winning methods leverage large-scale ensembles (6--11 models) as their core strategy, confirming that ensemble-based fusion remains a dominant approach for detection challenges with limited training data. The 1st place method additionally demonstrates that semi-supervised learning provides meaningful gains when labeled data is scarce, while the 3rd place method shows that annotation quality and domain-specific filtering (\S\ref{thermal-det:agaf}) can be competitive alternatives to semi-supervised learning. Across all submissions, small object detection remains the central difficulty: the extreme size distribution of thermal maritime objects (median navigational object area $\approx$154~px$^2$) pushes current detector architectures to their limits. Detailed method descriptions are provided in \S\ref{thermal-det:multi-arch-ssl}, \S\ref{thermal-det:deimv2-ensemble}, and \S\ref{thermal-det:agaf}.

\subsection{LaRS Panoptic Segmentation Challenge}
This challenge continues the LaRS panoptic benchmark and requires parsing USV-view scenes into \emph{stuff} classes and \emph{thing} instances. The \emph{stuff} classes include water, sky, and static obstacles, while the \emph{thing} classes include eight types of dynamic obstacles: boat/ship, buoy, row boat, swimmer, animal, paddle board, float and other. As in previous editions, strong performance requires balancing semantic consistency, instance-level detection, and robustness to small obstacles. 

\subsubsection{Evaluation Protocol}
Submitted methods are ranked based on panoptic quality (PQ) averaged over 11 classes and evaluated on 1,203 test images from the LaRS dataset. Different from standard evaluation, dynamic obstacle detections inside static obstacle regions are not additionally penalized as false positives~\cite{Zust2023LaRS}. For training, participants are allowed to use the LaRS training and validation sets, which consist of 2,605 and 198 labeled images respectively, as well as other datasets if disclosed. To provide additional insights, we factorize the panoptic quality into recognition quality (RQ) and segmentation quality (SQ)~\cite{Kirillov2019Panoptic}, as well as separately evaluate stuff (PQ$_\text{st}$) and thing (PQ$_\text{th}$) classes. 

\begin{table*}[ht]
\centering
\caption{Results for the USV-based Panoptic Segmentation Challenge measured in overall panoptic quallity (PQ) and separate for thing and stuff classes. Competing methods are compared to a baseline Mask2Former model~\cite{cheng2021mask2former} and the top-performing PanSR method~\cite{Zust2024PanSR}.}
\label{tab:usv-panoptic-overview}
\begin{tabular}{ccrrccccc}
\toprule
Place & Team & Method & Section & PQ & SQ & RQ &  PQ$_\text{st}$ & PQ$_\text{th}$ \\
\midrule
 $\textcolor{pink}{\bigstar}$ & MaCVi Team & PanSR (Swin-L) & - & 57.3 & 75.9 & 66.9 & 95.4 & 43.0 \\
\midrule
\mfirst{} & FER Zagreb & M2F-DINOv3 & \S\ref{panseg:m2f-dino} & \bm1{53.5} & \bm1{74.3} & \bm1{63.3} & \bm1{95.8} & \bm1{37.7} \\
\msecond{} & Fraunhofer IOSB - BIG & MaskDINOv3 & \S\ref{panseg:maskdino} & \bm2{48.3} & 72.4 & \bm2{57.7} & \bm3{94.9} & \bm2{30.9} \\
\mthird{} & KTSFOE & ThingSeg-DGCR & \S\ref{panseg:thingseg} & \bm3{42.6} & 71.0 & \bm3{51.1} & 91.8 & \bm3{24.2} \\
4th & PanopticMaritimeScenery & M2F-Mot3-SwinL-FPN & - & 42.2 & \bm2{73.5} & 48.7 & \bm2{95.3} & 22.3 \\
5th & NTNU & RF-DETR & - & 40.3 & 69.9 & 48.0 & 93.3 & 20.4 \\
6th & Dt deep-square & NewSR & - & 36.0 & \bm3{73.1} & 40.8 & 94.3 & 14.2 \\
\midrule 
- & MaCVi Team & M2F (Swin-B) & - & 41.4 & 75.2 & 47.1 & 92.5 &	22.3\\
\bottomrule
\end{tabular}
\end{table*}

\subsubsection{Submissions, Analysis and Trends}
The challenge received 26 submissions from 6 different teams. Table~\ref{tab:usv-panoptic-overview} shows the results for the top submission from each team, along with the performance of the state-of-the-art method PanSR~\cite{Zust2024PanSR} (out of competition) and our Mask2Former baseline with Swin-B backbone. Results from the remaining submissions are available on the public leaderboard on the MaCVi website. Further, we analyze the results with a focus on the top three performing teams, for which detailed technical reports are provided in the appendix~\ref{panseg:appendix}.

We observe that four of the six submitted methods outperform our baseline, while none surpass PanSR, the state-of-the-art method for maritime panoptic segmentation, which achieves 57.3 PQ on the LaRS test set. The top-performing submission, M2F-DINOv3, comes close with 53.5 PQ and even exceeds PanSR on stuff classes (PQ$_\text{st}$). However, the lower performance on thing classes underscores the importance of dynamic obstacle detection in maritime scenes. The second-ranked submission, MaskDINOv3, achieves 48.3 PQ, trailing the winner by 5.2 PQ points, while the third-ranked ThingSeg-DGCR reaches 42.6 PQ, trailing by 10.9 PQ points. Performance on thing classes appears to be the main factor differentiating the top three methods as well, with MaskDINOv3 and ThingSeg-DGCR falling behind M2F-DINOv3 by roughly 7 and 13 PQ$_\text{th}$ points, respectively.

\begin{figure}[h]
    \centering
    \includegraphics[width=\linewidth]{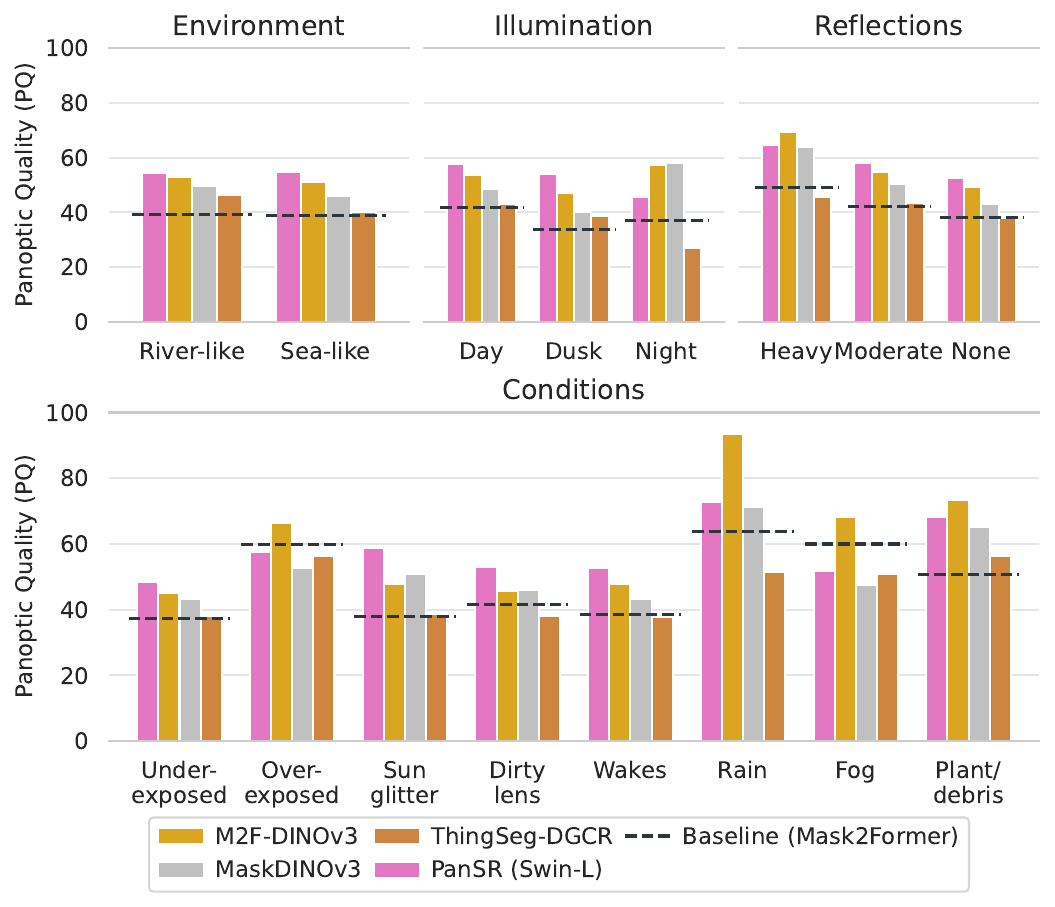}
    \caption{Panoptic quality stratified according to scene attributes.}
    \label{fig:panseg_scene_attr}
\end{figure}

\paragraph{Performance by scene attributes.} To further understand these differences, we now examine stratified performance according to different scene attributes including environment, illumination, reflections, and conditions. Figure~\ref{fig:panseg_scene_attr} shows that the global ranking is preserved across most of the scenarios. Interestingly, the top two submissions outperform the SOTA method PanSR in certain challenging scenarios, such as night illumination and heavy reflections. The winning M2F-DINOv3 further surpasses PanSR in overexposed scenes, as well as under rain, fog, and in the presence of plants or debris. These results indicate that large-scale backbone pretraining can enhance model robustness in difficult scenarios.

\begin{figure}[h]
    \centering
    \includegraphics[width=\linewidth]{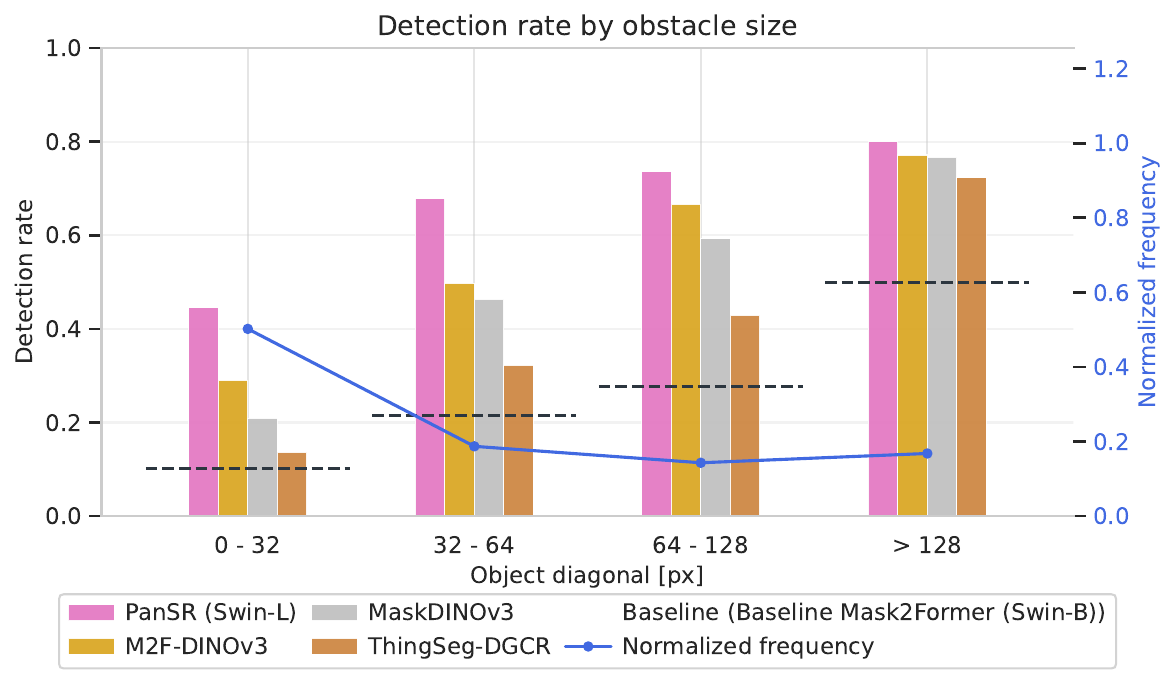}
    \caption{Dynamic obstacle detection rate stratified according to object size.}
    \label{fig:panseg_obj_size}
\end{figure}

\paragraph{Detection rate by obstacle size.}
Figure~\ref{fig:panseg_obj_size} shows the detection rates of dynamic obstacles as a function of object size. It also shows the normalized object frequency for each size bin (blue). Detection rates are lowest for the smallest objects (first bin), which also contains the largest number of instances. This is where the gap between the submitted methods and the SOTA PanSR is largest, highlighting the impact of PanSR's object-centric proposal head and training scheme, that addresses the oversegmentation and mask drifting in inference typical for Mask-Dino-based models, and improves small object detection and dense scenarios.

\paragraph{Qualitative results.}
Figure~\ref{fig:panseg_qual} presents qualitative results on three LaRS test scenes for the top-performing methods. The largest differences appear in the third scene, which contains many small objects. Even PanSR fails to detect all objects labeled in the ground truth, and the detection rate is lower for the other methods. Interestingly, MaskDINOv3 detects more objects than M2F-DINOv3 but struggles to segment the sea, highlighting differences in how Mask2Former and MaskDINO generate mask proposals.
\newcommand{\imgwidth}{0.163\textwidth}

\begin{figure*}[h]
\centering
\setlength{\tabcolsep}{1pt} 
\begin{tabular}{c*{6}{c}}
Image & GT & PanSR~\cite{Zust2024PanSR} & \mfirst{}M2F-DINOv3 & \msecond{}MaskDINOv3 & \mthird{}ThingSeg-DGCR \\

\includegraphics[width=\imgwidth]{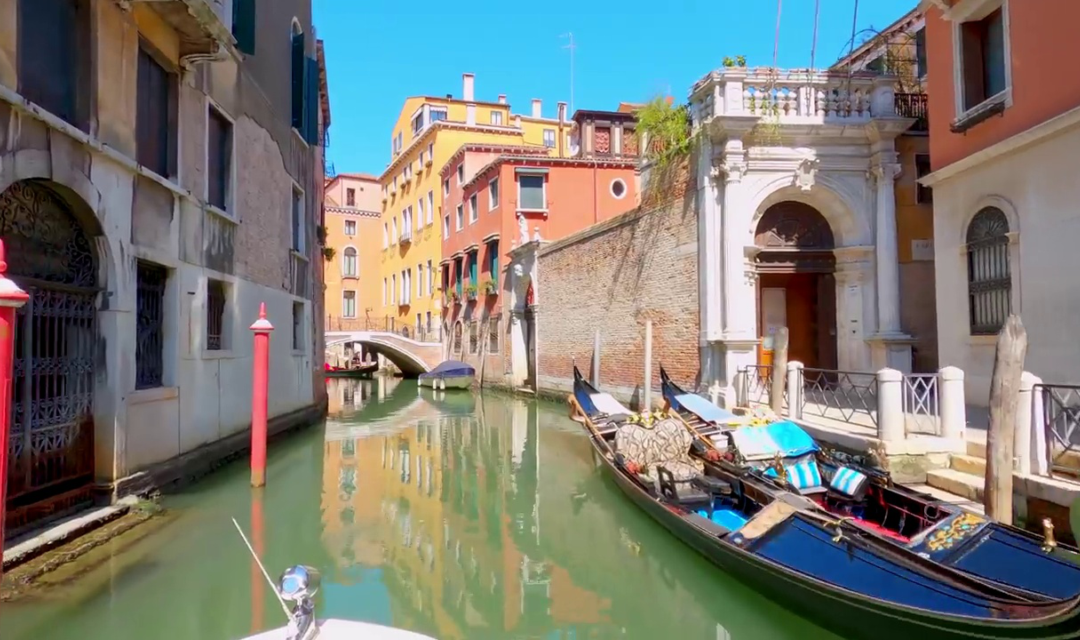} &
\includegraphics[width=\imgwidth]{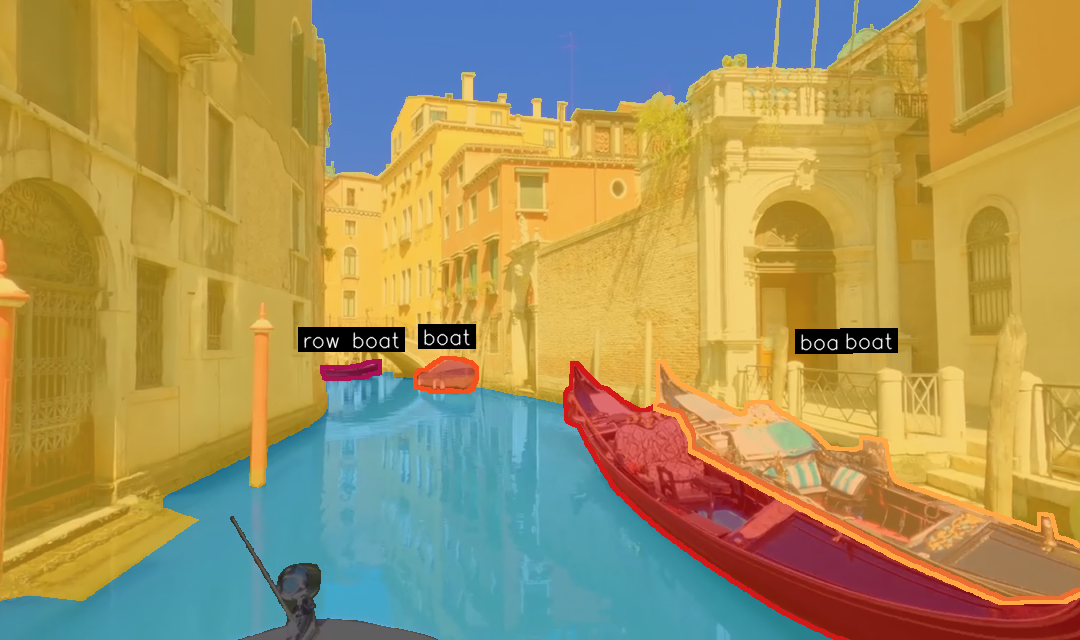} &
\includegraphics[width=\imgwidth]{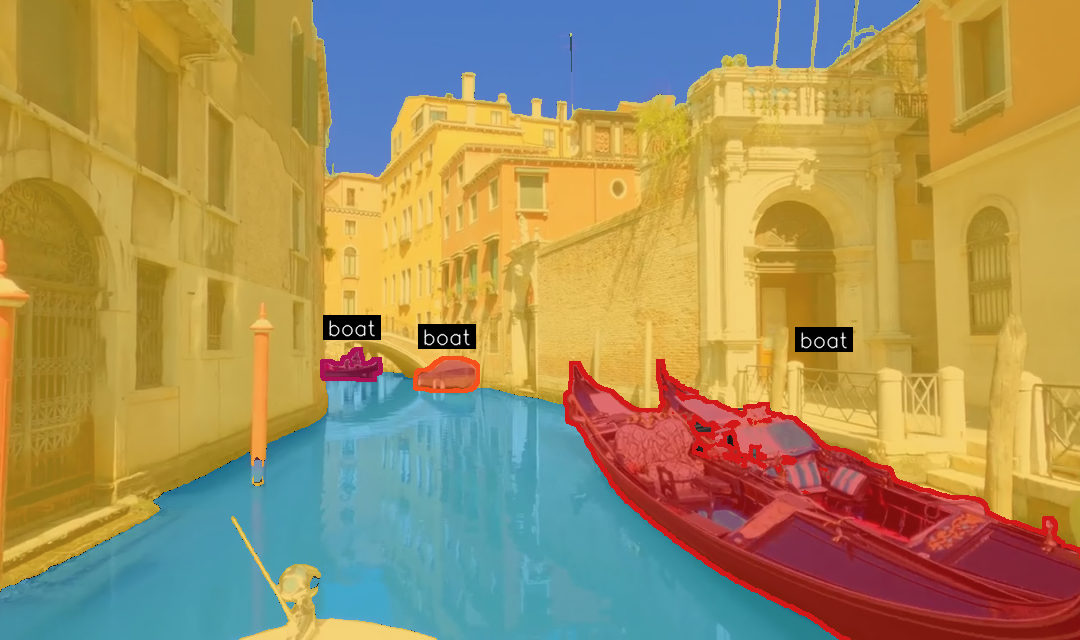} &
\includegraphics[width=\imgwidth]{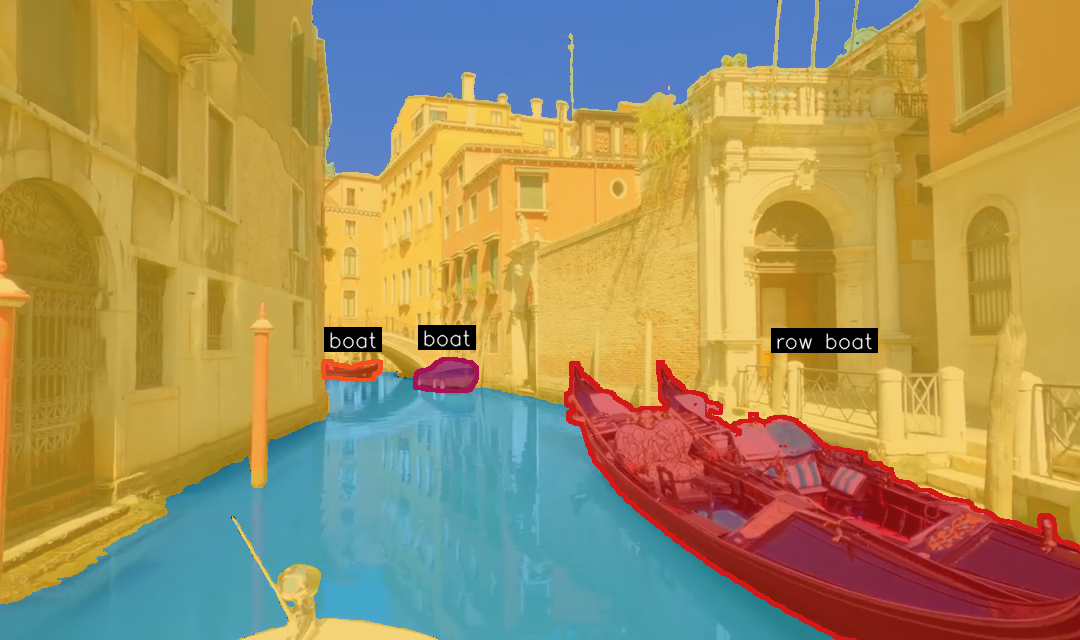} &
\includegraphics[width=\imgwidth]{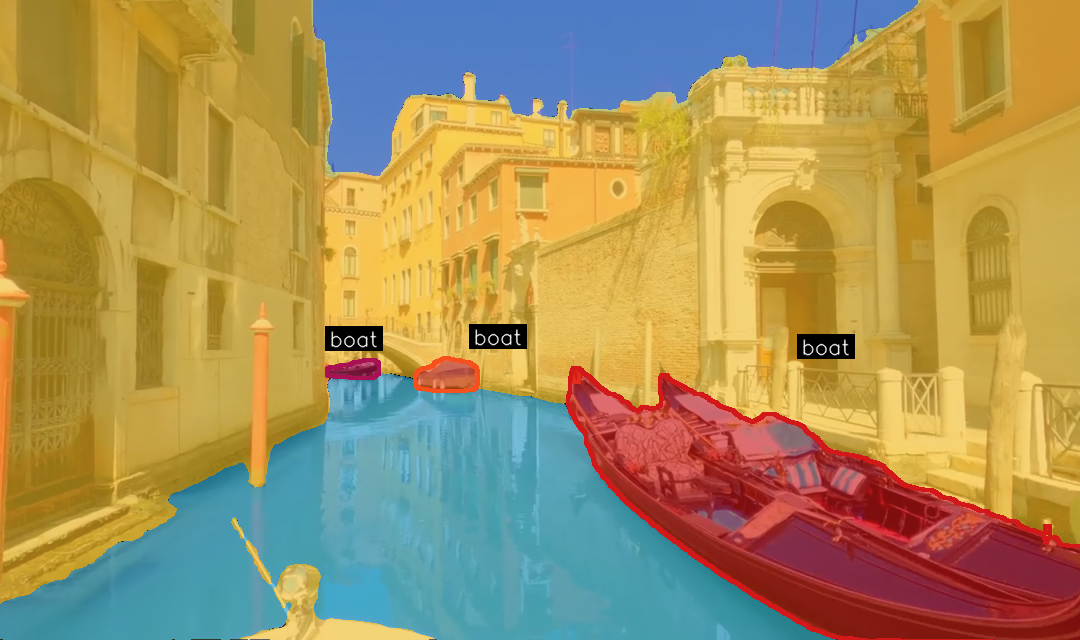} &
\includegraphics[width=\imgwidth]{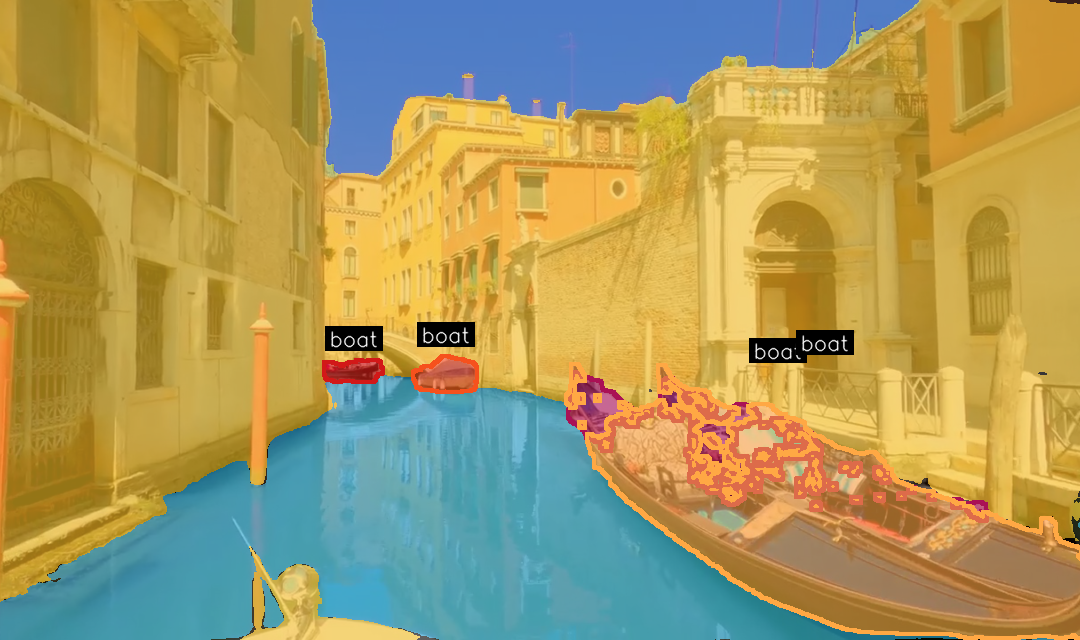} \\[-0.2em]

\includegraphics[width=\imgwidth]{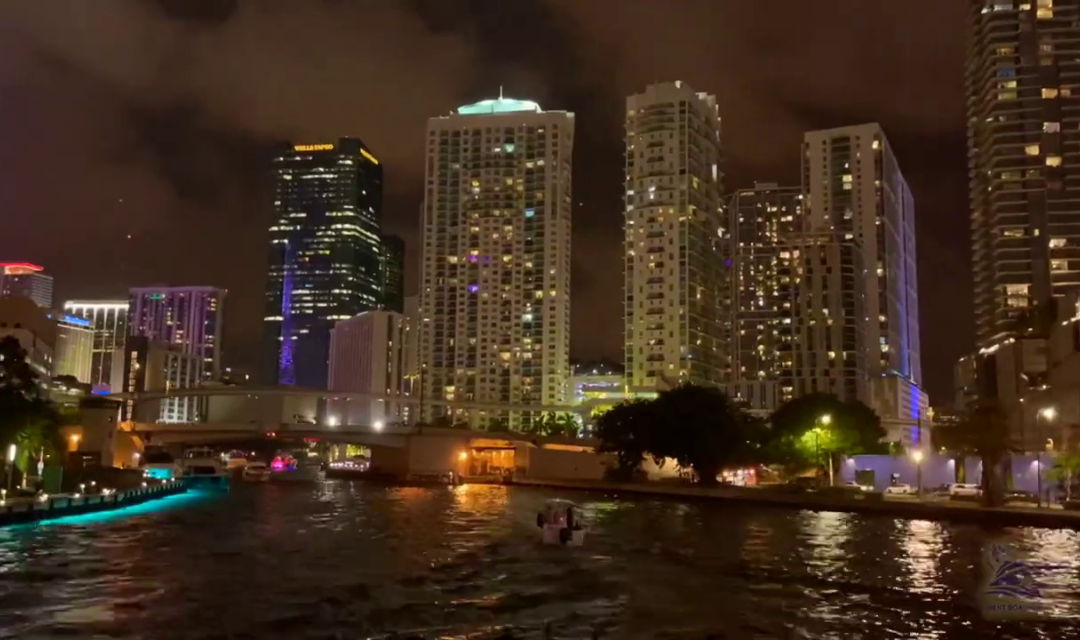} &
\includegraphics[width=\imgwidth]{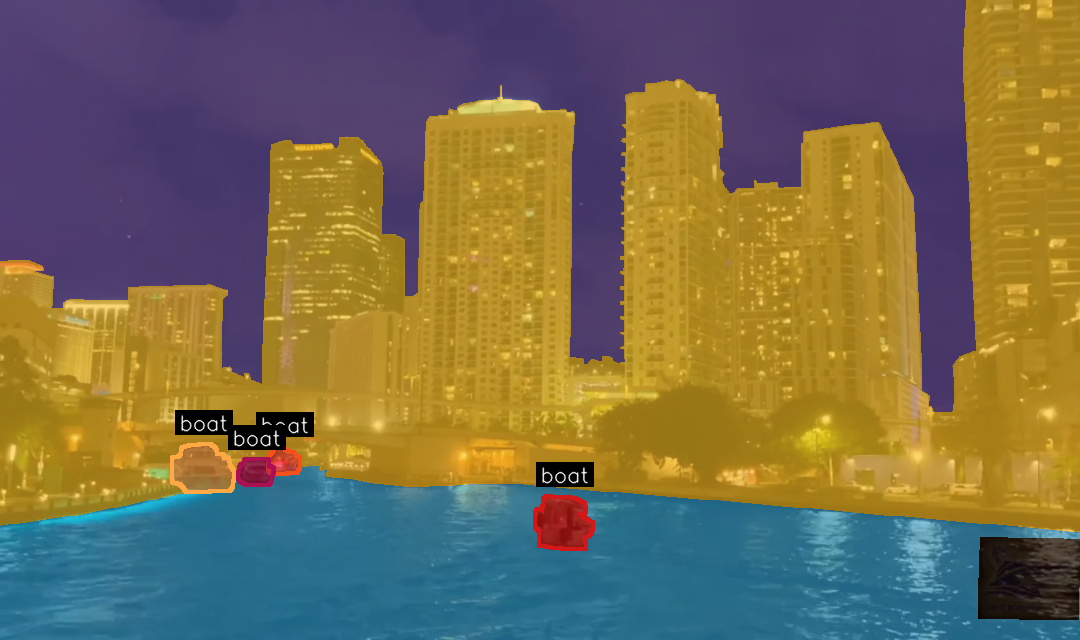} &
\includegraphics[width=\imgwidth]{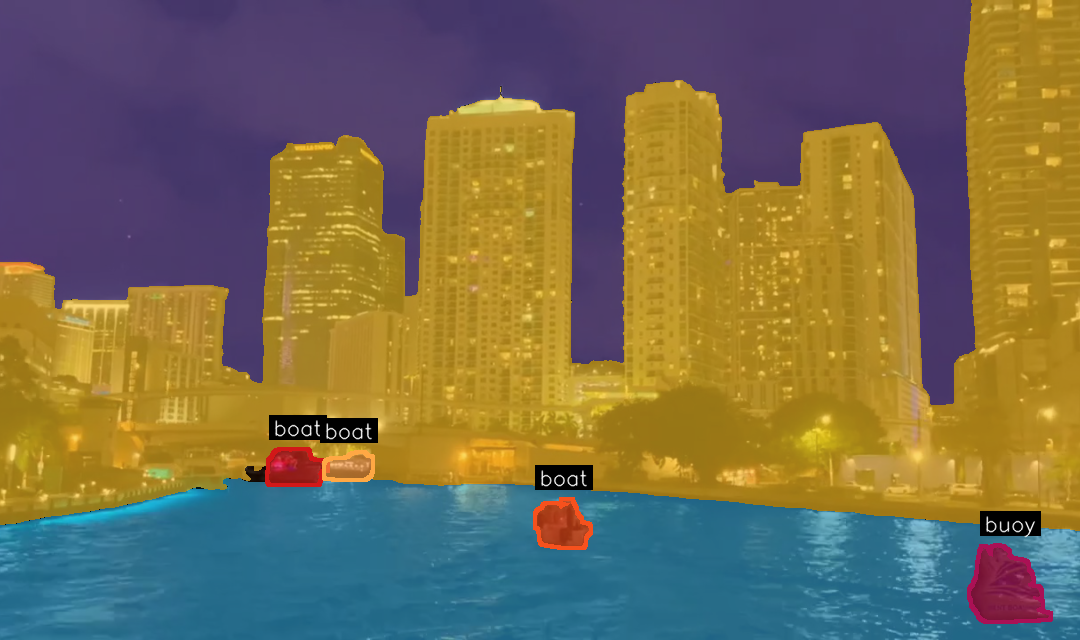} &
\includegraphics[width=\imgwidth]{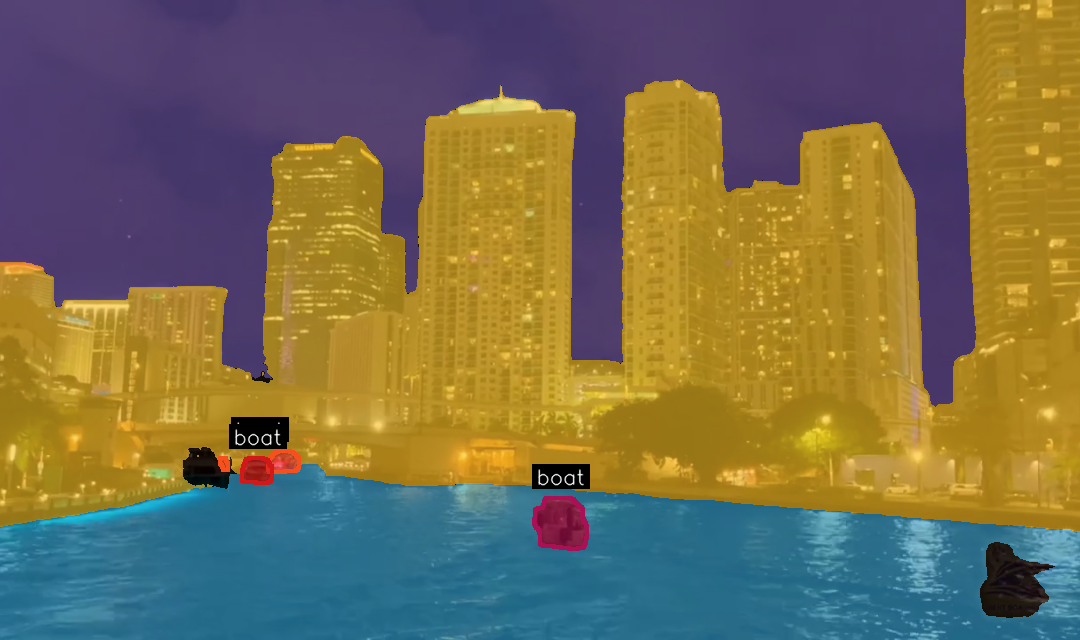} &
\includegraphics[width=\imgwidth]{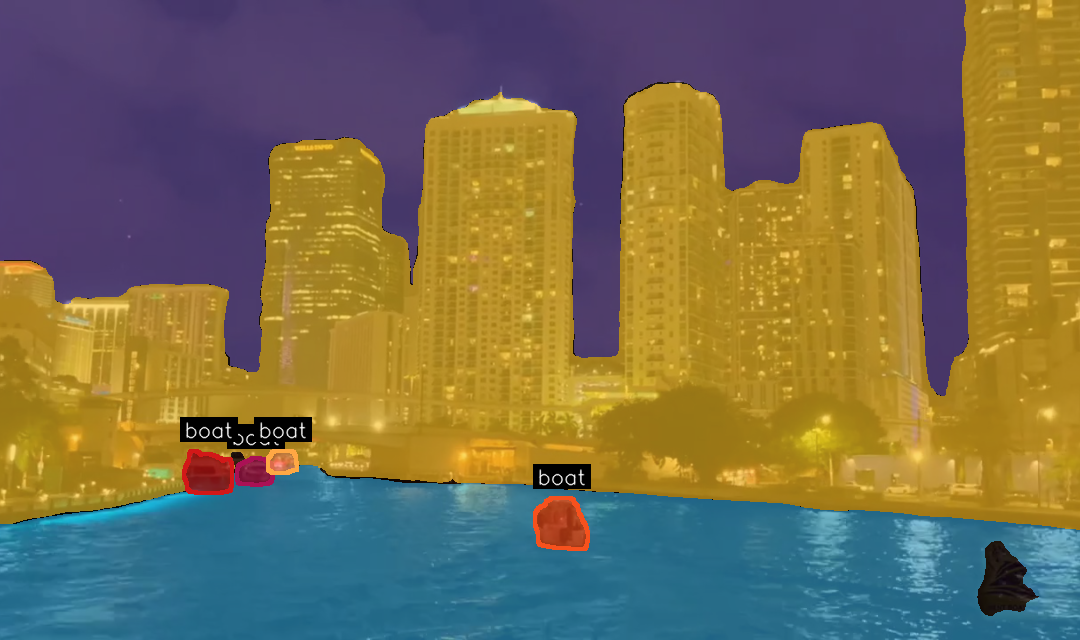} &
\includegraphics[width=\imgwidth]{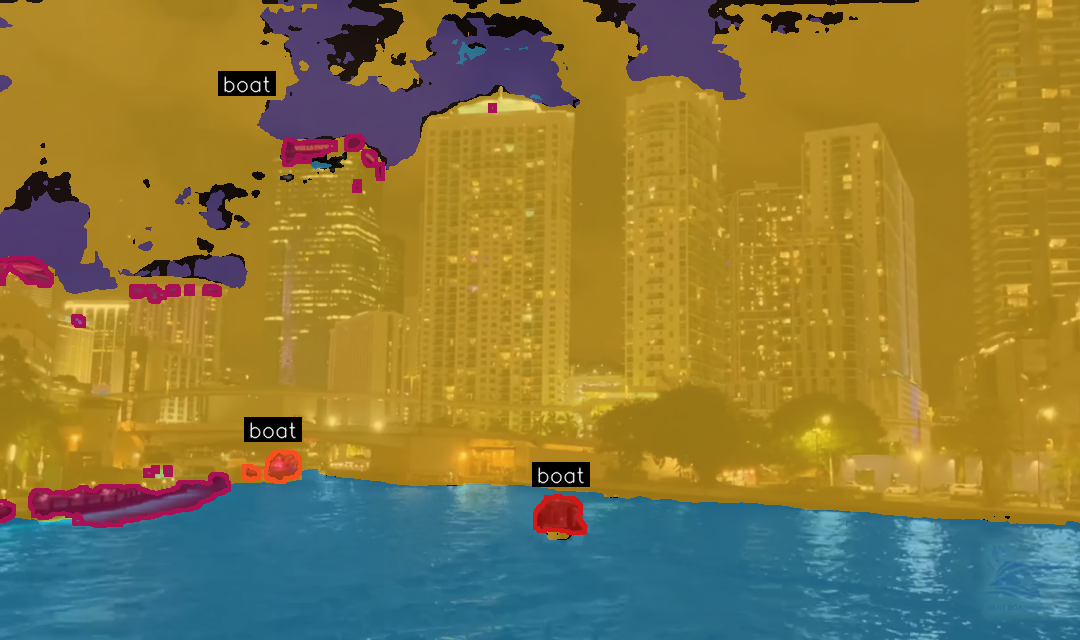} \\[-0.2em]

\includegraphics[width=\imgwidth]{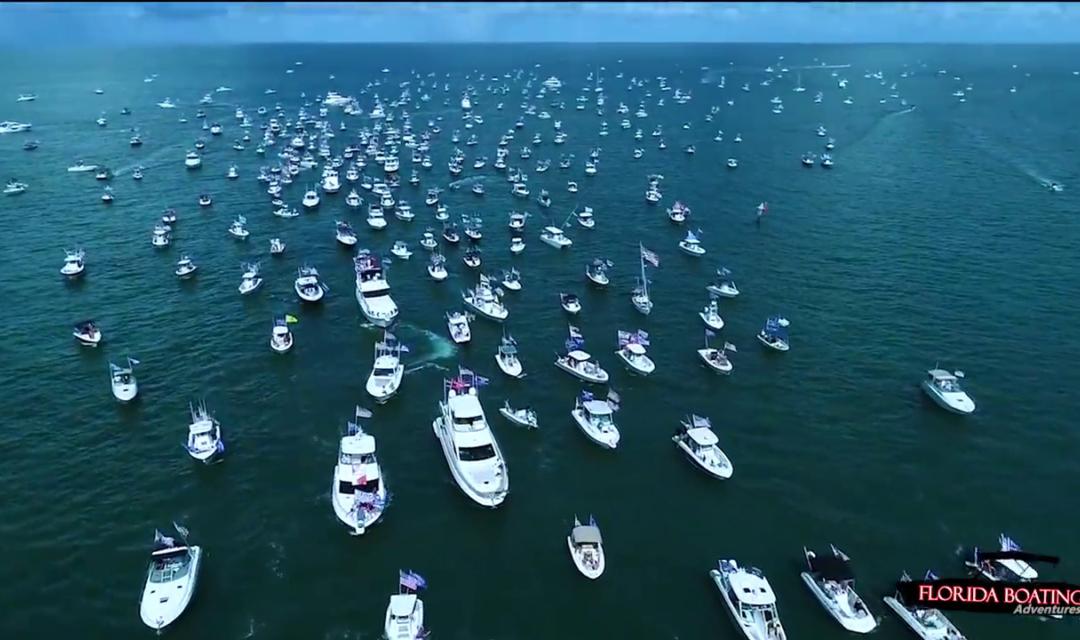} &
\includegraphics[width=\imgwidth]{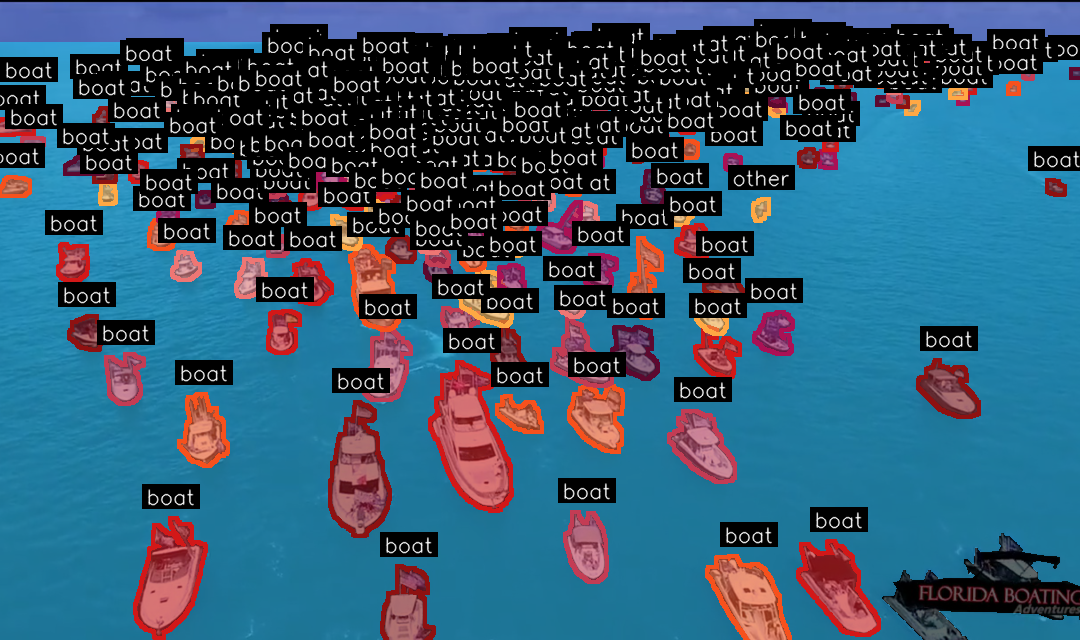} &
\includegraphics[width=\imgwidth]{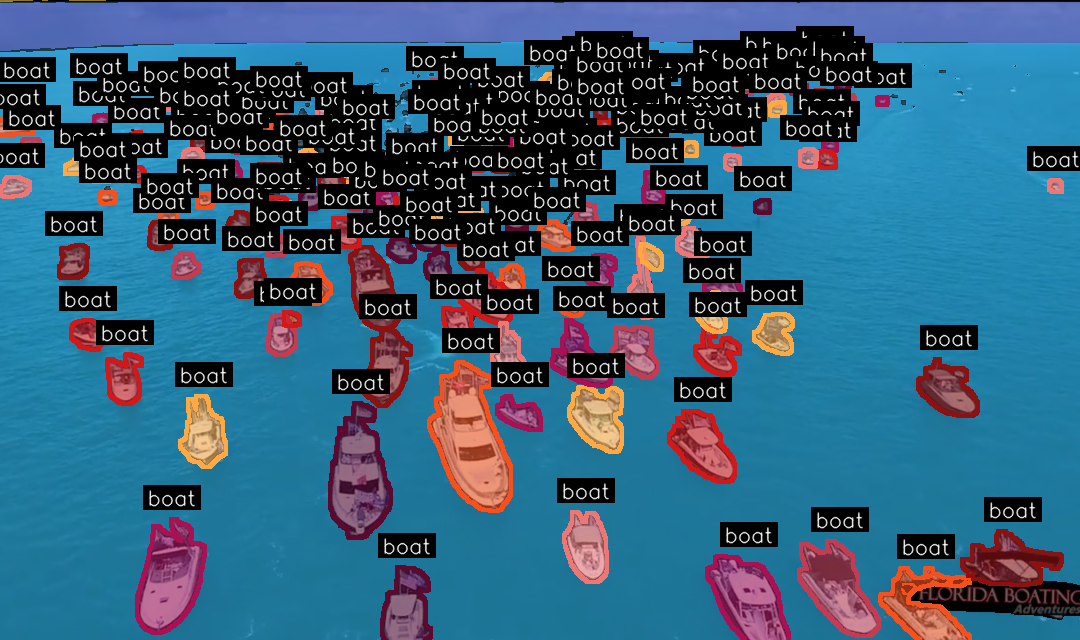} &
\includegraphics[width=\imgwidth]{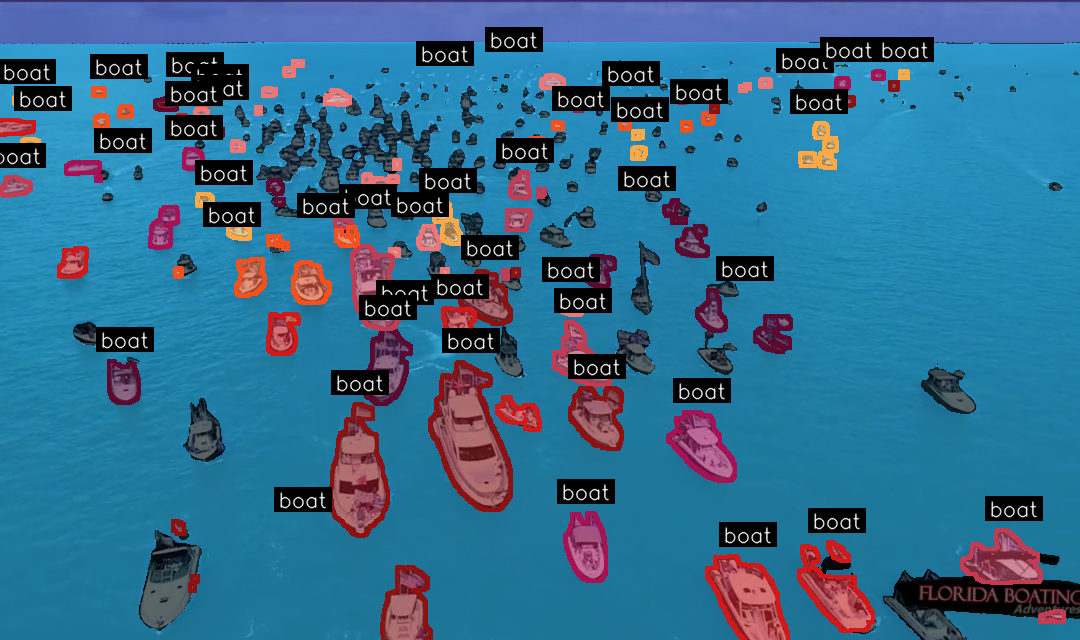} &
\includegraphics[width=\imgwidth]{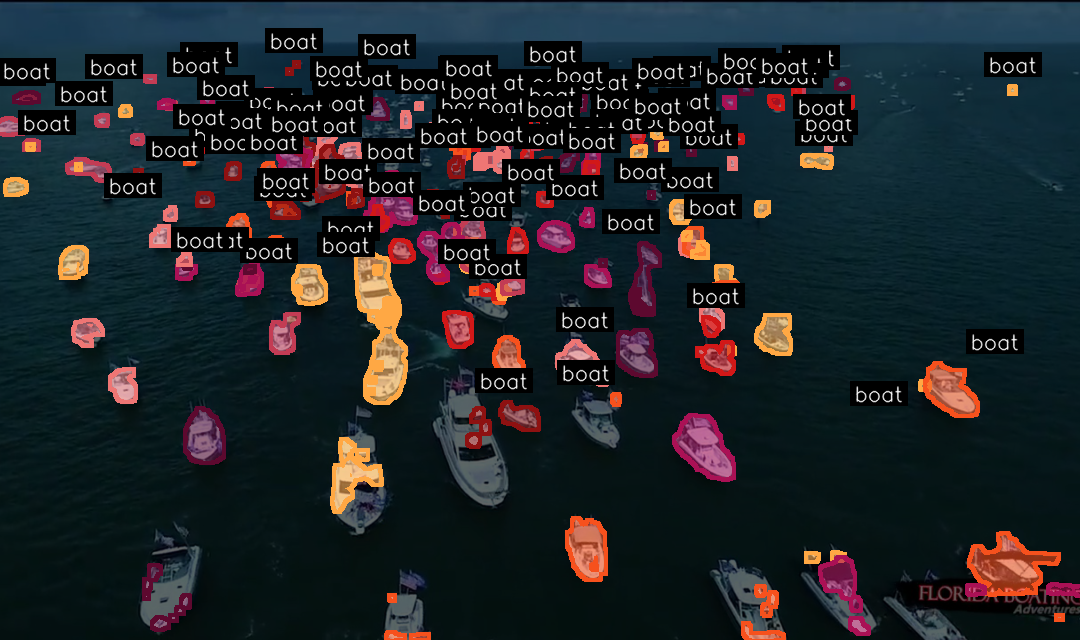} &
\includegraphics[width=\imgwidth]{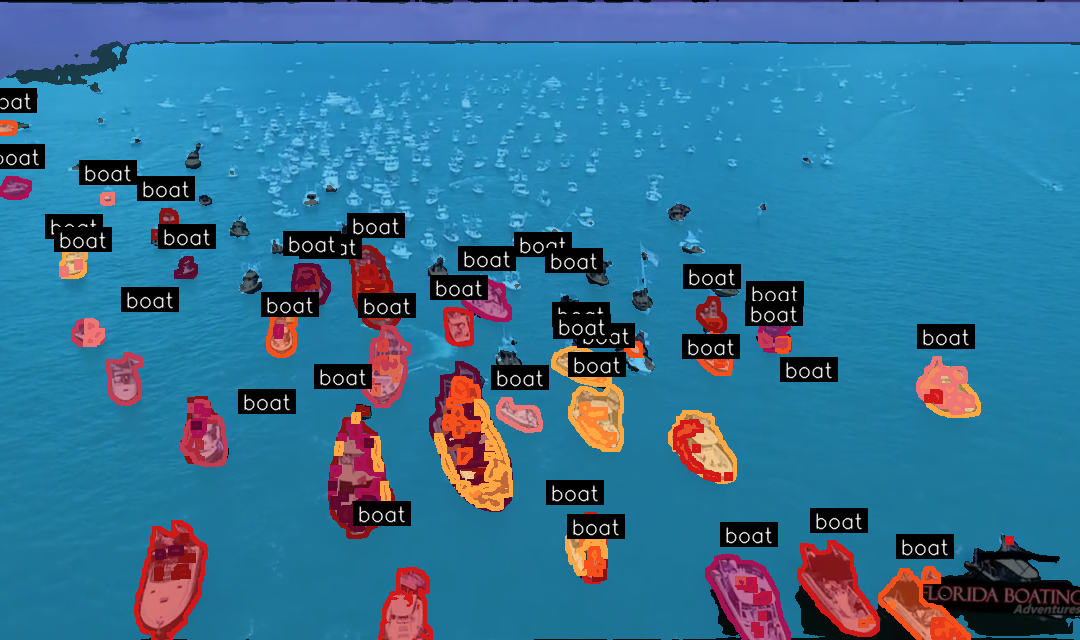} \\

\end{tabular}
\caption{Qualitative results of the top-performing methods in the LaRS panoptic segmentation challenge.}
\label{fig:panseg_qual}
\end{figure*}

\subsubsection{Discussion and Challenge Winners}

The winners of the LaRS Panoptic Segmentation challenge are listed in Table \ref{tab:usv-panoptic-overview}. All three methods outperform our baseline and show notable improvements in both stuff and thing segmentation. The performance of the winning method is comparable to the winner of the previous challenge, yet it still falls short of our baseline specialized for maritime panoptic segmentation. The top two methods highlight the strong trend of using mask-transformer meta-architectures with large-scale, self-supervised backbones. While self-supervised pretraining improves robustness under challenging conditions such as rain and fog, detecting small objects in crowded scenes remains challenging. Further gains may come from combining large-scale self-supervised pretraining~\cite{simeoni2025dinov3} with targeted small-object detection strategies~\cite{Zust2024PanSR}.

\subsection{Embedded Semantic Segmentation Challenge}
Recent maritime obstacle segmentation methods often rely on expensive, power-hungry hardware, making them unsuitable for small, energy-constrained USVs. Building on previous editions~\cite{Kiefer_2024_WACV, Kiefer_2025_WACV}, this challenge targets methods that balance segmentation quality and real-time efficiency and evaluates them on a real-world Luxonis embedded device OAK4~\cite{luxonis_oak4d}.

\subsubsection{Evaluation Protocol}

The Embedded Semantic Segmentation Challenge follows the LaRS evaluation protocol~\cite{Zust2023LaRS} with additional deployment constraints for embedded execution. Methods predict per-pixel labels for water, sky, and obstacles and are evaluated with navigation-oriented rather than conventional segmentation metrics.

Following LaRS, static obstacle detection is measured by water-edge accuracy ($\mu$), while dynamic obstacle detection is summarized by F1. To balance dynamic obstacle detection and segmentation quality, the primary ranking metric is the combined quality score
\begin{equation}
Q = \mathrm{F1} \cdot \mathrm{mIoU}.
\end{equation}

Because submissions must run on the target embedded platform, models must be exportable to ONNX with a static computation graph, use only supported operations, accept a single $768 \times 384$ image normalized with ImageNet~\cite{deng2009imagenet} statistics, and achieve at least 30 FPS on the target device.

For server-side evaluation, submitted models are quantized to INT8 using the LaRS validation split, compiled for device execution, and evaluated after standard resizing and padding, with outputs resized back to the original resolution before scoring. Final rankings are computed with the LaRS metrics above, and average embedded throughput is reported alongside accuracy.

\subsubsection{Submissions, Analysis and Trends}

\begin{table*}[t]
\centering
\caption{Overview of the submissions for the USV-based Embedded Obstacle Segmentation challenge. For comparison, last year's winning method is included inline in gray and is not assigned placements. The best results among current non-gray entries are denoted in bold.}
\label{tab:embedded-seg-overview-cvpr26}
\begin{tabular}{cllcccccccc}
\toprule
Place & Institution & Method & Section & FPS & \textbf{Q}$\downarrow$ & $\mu$ & Pr & Re & F1 & mIoU \\
\midrule
\color{gray} $*$ & \color{gray} DLMU & \color{gray} RSOS-Net & \color{gray} \cite{Kiefer_2025_WACV} & \color{gray}\textcolor{gray}{85.1} & \color{gray}64.2 & \color{gray}72.5 & \color{gray}66.4 & \color{gray}67.9 & \color{gray}67.1 & \color{gray}95.7 \\
\mfirst{} & Independent Researcher & DSOS-Net & \S\ref{embseg:dsos-net} & \textcolor{teal}{66.5} & \textbf{61.9} & \textbf{68.3} & \textbf{62.6} & 69.7 & \textbf{66.0} & 93.8 \\
\msecond{} & Independent / K-State & PIDNet-S & \S\ref{embseg:pidnet-s} & \textcolor{teal}{67.6} & 52.1 & 67.8 & 54.1 & 56.4 & 55.2 & \textbf{94.4} \\
\mthird{} & Colorado School of Mines & RSOS\_R50 & \S\ref{embseg:rsos-net} & \textcolor{teal}{\textbf{96.3}} & 46.4 & 61.5 & 41.6 & 67.0 & 51.3 & 90.3 \\
- & Colorado School of Mines & YXHR\_medium & \S\ref{embseg:rsos-net} & \textcolor{teal}{64.9} & 45.3 & 64.3 & 42.6 & 64.3 & 51.3 & 88.4 \\
5th & DT Deep-Square & newlight & - & \textcolor{teal}{130.1} & 39.4 & 56.3 & 32.1 & \textbf{75.8} & 45.1 & 87.3 \\
6th & UCLA & EdgeModel & - & \textcolor{teal}{32.4} & 18.5 & 54.9 & 13.9 & 59.9 & 22.5 & 82.3 \\

\bottomrule
\end{tabular}
\end{table*}

We received 20 submissions from 5 teams. Final rankings assign one placement per team using the best-ranked submission when a team enters multiple models. Table~\ref{tab:embedded-seg-overview-cvpr26} also includes last year's winning method~\cite{Kiefer_2025_WACV} for comparison, but that entry is not part of the ranking.

The top three methods highlight complementary strategies for embedded maritime segmentation. DSOS-Net (\S\ref{embseg:dsos-net}) leads overall by combining a DINOv3-pretrained ConvNeXt backbone~\cite{simeoni2025dinov3,liu2022convnet} with a lightweight RSOS-Net-style decoder~\cite{wang2026rsosnet} and a two-stage training schedule. PIDNet-S (\S\ref{embseg:pidnet-s}) emphasizes quantization-friendly real-time deployment and Copy-Paste obstacle augmentation for small and rare obstacles. RSOS\_R50 (\S\ref{embseg:rsos-net}) revisits RSOS-Net in PyTorch with a ResNet-50 backbone and lightweight multi-scale context aggregation.

Compared to last year's reference model, the top-ranked methods do not surpass the best $Q$ or F1 scores, but they remain above the embedded speed threshold. Among this year's ranked entries, DSOS-Net achieves the best quality score and F1, PIDNet-S the best mIoU, and RSOS\_R50 the highest throughput of the top three, highlighting the trade-off between quality and throughput.

\begin{figure}[t]
\centering
\includegraphics[width=0.98\linewidth]{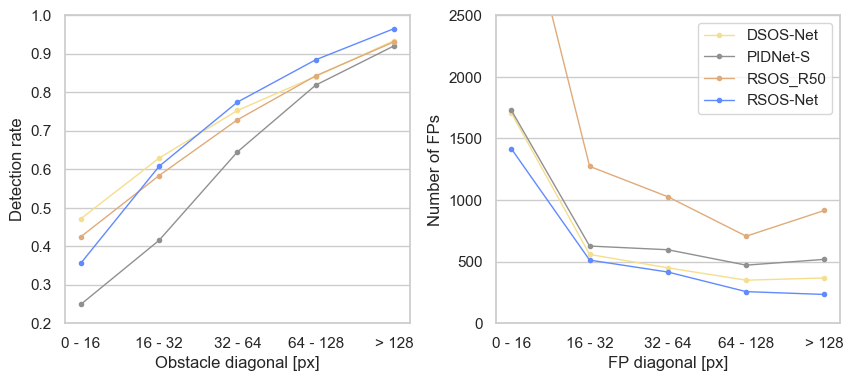}
\caption{Detection rate and FP as a function of obstacle size for embedded segmentation methods. Only the three winning and the best-performing method from last year's challenge are shown.}
\label{fig:embseg-analysis-size}
\end{figure}

\begin{figure}[t]
\centering
\includegraphics[width=\linewidth]{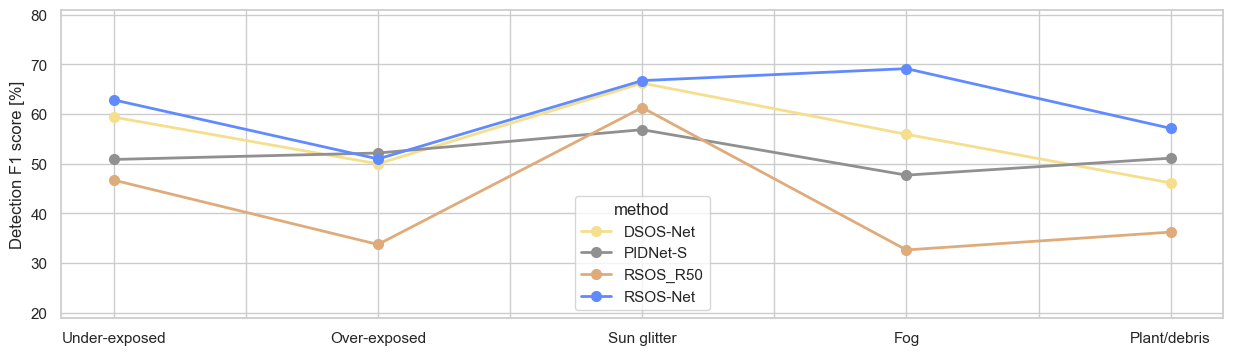}
\caption{Detection F1 under challenging visual conditions for embedded segmentation methods. Only the three winning and the best-performing method from last year's challenge are shown.}
\label{fig:embseg-analysis-conditions}
\end{figure}

\begin{figure}[t]
\centering
\includegraphics[width=0.98\linewidth]{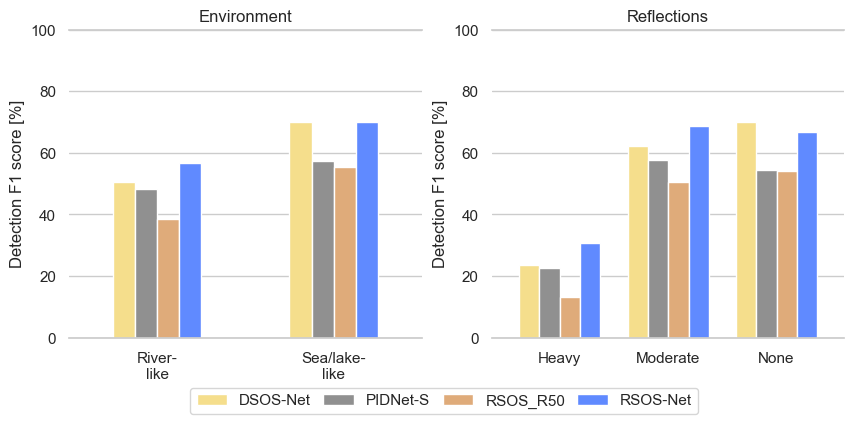}
\caption{Detection F1 across scene environments and reflection levels for embedded segmentation methods showing the three winning and the best-performing method from last year's challenge.}
\label{fig:embseg-analysis-environment}
\end{figure}

Across obstacle sizes in Figure~\ref{fig:embseg-analysis-size}, DSOS-Net is strongest on the smallest obstacles, with RSOS\_R50 also competitive, whereas last year's RSOS-Net remains strongest on medium and large obstacles. RSOS-Net and DSOS-Net also produce the fewest false positives, while RSOS\_R50 generates more small-object false alarms. Under scene conditions and environments (Figures~\ref{fig:embseg-analysis-conditions} and \ref{fig:embseg-analysis-environment}), last year's RSOS-Net remains strongest in many difficult settings, DSOS-Net is the best current entry overall and strongest when reflections are absent, and PIDNet-S is generally more stable than RSOS\_R50 but less accurate. Overall, DSOS-Net is the strongest ranked method this year, while last year's RSOS-Net remains a strong reference.

\begin{table*}[!b]
\centering
\caption{Results for the USV-based Multimodal Segmentation Challenge. We report $M$, validation/test mIoU, and dynamic-obstacle IoU.}
\label{tab:usv-multimodal-overview}
\resizebox{\linewidth}{!}{%
\begin{tabular}{ccrrccccc}
	\toprule
	Place & Team & Method & Section & M & mIoU (\emph{val}) & IoU$_{do}$ (\emph{val}) &  mIoU (\emph{test}) & IoU$_{do}$ (\emph{test}) \\

	\midrule
	$\textcolor{pink}{\bigstar}$ & MaCVi Team & SWIN-B + Mask2Former & - & 83.4 & 90.8 & 70.1 & 76.0 & 37.0 \\
	\midrule
	\mfirst{} & GIST AI LAB & GatedMemorySAM & \S\ref{mmseg:GatedMemorySAM} & \bm1{82.1} & \bm1{93.3} & \bm1{78.8} & \bm1{70.9} & \bm1{35.4} \\
	\msecond{} & Xidian University & Adapted MFNet-SAM-LoRA & \S\ref{mmseg:MFNet} & \bm2{64.2} & \bm2{90.9} & \bm2{70.3} & \bm2{37.6} & \bm3{7.4} \\
	\mthird{} & Fraunhofer IOSB & Modified DustNet & \S\ref{mmseg:DustNet} & \bm3{62.8} & \bm3{87.9} & \bm3{60.7} & \bm2{37.6} & \bm2{13.5} \\
	\bottomrule
\end{tabular}
}
\end{table*}

\subsubsection{Discussion and Challenge Winners}

Table~\ref{tab:embedded-seg-overview-cvpr26} lists the winners of the Embedded Semantic Segmentation challenge. DSOS-Net leads overall, PIDNet-S prioritizes quantization-friendly real-time deployment, and RSOS\_R50 remains competitive when throughput is prioritized.

\subsection{Multimodal Semantic Segmentation Challenge}
This track studies segmentation with synchronized RGB, thermal, and LiDAR inputs on MULTIAQUA. Methods predict four classes---\emph{static obstacle}, \emph{dynamic obstacle}, \emph{water}, and \emph{sky}---and must remain reliable when one or more modalities are degraded or missing.

\subsubsection{Evaluation Protocol}

Participants submitted predictions for the labeled validation/daytime split and the hidden test/nighttime split. The main challenge is two-fold: methods must exploit auxiliary modalities such as thermal and LiDAR while still relying on RGB when it is informative. Ranking is based on $M$, the mean of the two mIoU scores; per-split mIoU and dynamic-obstacle IoU are also reported to capture nighttime robustness and safety-critical obstacle detection. 

\subsubsection{Submissions, Analysis and Trends}
The challenge received 36 submissions from 3 teams, with only each team's best entry used for the final ranking. Technical reports for the top-performing methods are provided in Appendix~\ref{mmseg:appendix}. Two of the submitted methods use strong pretrained backbones from the SAM family, giving them a strong start with the RGB input, while potentially also helping with interpreting other modalities. As shown by the winner, GatedMemorySAM, explicitly modeling difficult visibility conditions is a necessary part of training robust multimodal models.

\subsubsection{Discussion and Challenge Winners}

The winners of the Multimodal Semantic Segmentation challenge are listed in Table~\ref{tab:usv-multimodal-overview}. GatedMemorySAM performs best overall and even outperforms the MaCVi reference on the validation split. The other two methods remain competitive on daytime data but degrade markedly at night, showing that robust cross-modal fusion under degraded sensing remains difficult.

\section{Conclusion}

This summary presents the MaCVi 2026 challenges and we see a trend towards maritime perception methods that transfer across tasks, sensing conditions, and deployment constraints. Future iterations will introduce more generalist challenges.

\vspace{.2cm}
\noindent\textbf{Acknowledgments.}
We thank all participating teams, contributors, and organizers supporting the MaCVi initiative. We also thank \href{https://www.catskill.ai/}{catskill GmbH} for sponsoring the RTX 5080 GPU prize, \href{https://www.luxonis.com/}{Luxonis} for sponsoring the embedded challenge prize, and \href{https://getalookout.com/}{LOOKOUT} for data support.


\newpage

\appendix
\addcontentsline{toc}{section}{Submitted Methods}
\section*{Appendix - Meta Challenge + Tech Reports}

\subsection*{Generalist Meta Challenge}

The Generalist Meta Challenge evaluates whether a method remains competitive across multiple maritime tasks rather than excelling on only one benchmark. Teams must participate in at least two active challenge tracks and submit a short report (maximum four double-column pages, excluding references) that justifies why the method should be considered generalist and lists all evaluated tracks.

\paragraph{Evaluation protocol and metrics.} Let there be $M$ challenge leaderboards, with leaderboard $i$ containing $N_i$ ranked entries. For each model $j$, the per-leaderboard rank is defined as
\begin{equation}
r_{ij}=\begin{cases}
\text{observed rank of }j, & \ j \in leaderboard\ i\\
N_i+1, & \ j \notin leaderboard\ i
\end{cases}
\end{equation}
Each rank is converted to a normalized score in $[0,1]$ using
\begin{equation}
s_{ij}=\max\left(0,1-\frac{r_{ij}-1}{N_i-1}\right),
\end{equation}
so rank $1$ maps to $1$, the last listed rank $N_i$ maps to $0$, and missing entries ($N_i+1$) also map to $0$. The final metric is the \emph{Consistency Score}, computed as the arithmetic mean across leaderboards:
\begin{equation}
C_j=\frac{1}{M}\sum_{i=1}^{M}s_{ij}.
\end{equation}
This protocol makes scores comparable across heterogeneous leaderboards and penalizes selective participation, because non-submitted tracks contribute zero to the final average.

\paragraph{Participation.} As this is a newly introduced challenge in the current MaCVi edition, no teams submitted eligible entries for the Generalist Meta Challenge. Accordingly, no official rankings, consistency scores, or winning methods are reported for this track in this version of the workshop summary.

\newpage

\subsection*{Technical Reports}

\section{Vision-to-Chart Data Association Challenge}
\label{visionmap:submissions}

\subsection{\texorpdfstring{\protect\mfirst{}}{(1st)} Skyline-Aware ROI-Calibrated Buoy Association}
\label{visionmap:skyline-roi}
\noindent
\emph{Wonwoo Jo, Hansol Kim, Hyewon Chun, Sangmun Lee, Jeeyeon Jeon}\\
\emph{HD Korea Shipbuilding \& Offshore Engineering Co., Ltd.}\\

\noindent
The winning submission uses a staged pipeline that explicitly injects geometric structure before final buoy--query assignment. The method first estimates the skyline to compensate vessel roll, then projects each chart query into a query-conditioned image ROI using learned distance- and bearing-dependent calibration curves. Buoy candidates are generated with RF-DETR 2XL from both the full image and ROI crops, matched to chart queries with Hungarian assignment, and filtered with a gradient-boosted calibrator trained on assignment features. According to the submitted report, this design improves the overall validation score from 0.317 for the organizer baseline to 0.885 on the validation and from X to 0.765 on the test split in the final packaged system, and achieved the top leaderboard performance on both splits.

\subsection{\texorpdfstring{\protect\msecond{}}{(2nd)} Learned World-to-Image Projection for Query-Conditioned DETR}
\label{visionmap:querymlp}
\noindent
\emph{Borja Carrillo-Perez}\\
\emph{Arquimea Research Center}\\

\noindent
The second-place method extends the organizer's DETR-style baseline with a dedicated QueryMLP that predicts the buoy's waterline contact point directly in image coordinates from chart measurements and IMU orientation. The predicted pixel location is appended to the query embedding, reducing the geometric reasoning burden on the transformer decoder and providing a stronger spatial prior for association. The report emphasizes the use of normalized distance, inverse distance, bearing, pitch, roll, and heading as inputs to the projection MLP, followed by a standard DETR training pipeline with score-bias calibration at inference. This lightweight modification achieved an overall private-test score of 0.7386 and ranked second on the official leaderboard.

\subsection{\texorpdfstring{\protect\mthird{}}{(3rd)} Dynamic Chart-Derived Queries with DEIMv2}
\label{visionmap:deimv2-dynamic}
\noindent
\emph{Yusi Cao, Jiahui Wang, Lingling Li, Xu Liu, Licheng Jiao}\\

\noindent
The third-place report builds on the DEIMv2 detection framework with a DINOv3 ViT-Tiny backbone and adapts it to the vision-to-chart setting through dynamic chart-derived queries. Each chart marker is encoded by a lightweight MLP into a hidden embedding and processed jointly with multi-scale image features in a decoder that supports a variable number of valid queries via masking. The network predicts both a visibility confidence and normalized bounding-box coordinates for each query. The submitted model contains approximately 8.37M parameters, is trained for 120 epochs with AdamW and cosine decay, and uses horizontal flips together with query noise augmentation on distance and bearing. The reported validation performance was 0.409 precision, 0.385 recall, 0.397 F1, and 0.363 mIoU, corresponding to third place in the challenge ranking.

\subsection{(4th) IMU-Conditioned Query DETR for Maritime Buoy Detection}
\label{visionmap:imu-detr}
\noindent
\emph{Vinayak Nageli$^1$, Arshad Jamal$^2$, Rama Krishna S. Gorthi$^1$}\\
\emph{$^1$Indian Institute of Technology Tirupati, $^2$Centre for Artificial Intelligence and Robotics (CAIR), DRDO}\\

\noindent
The fourth-place submission replaces the standard learned object queries of the baseline with geometry-aware query embeddings derived from chart distance, chart bearing, and an encoded IMU state vector. The model uses a ResNet-50 backbone and a transformer detector whose query representation is produced by a small MLP operating on distance, sine/cosine bearing encoding, and IMU context. To improve localization of small buoys, the authors additionally introduce a weighted horizontal-center loss, motivated by the importance of accurate lateral placement in this task. Training is performed in two stages with AdamW and cosine annealing, first with BCE+L1+GIoU losses and then with an L1-focused continuation stage. The report highlights improved convergence from navigational priors. However, the submission reports a validation overall score of 0.2147, which could not be reproduced by the challenge organizers.


\section{Thermal Object Detection Challenge}

\subsection{\texorpdfstring{\protect\mfirst{}}{(1st)} Multi-Architecture Ensemble with Semi-Supervised Learning}
\label{thermal-det:multi-arch-ssl}
\noindent
\emph{Tze-Hsiang Tang}\\
\texttt{seantangth@gmail.com}\\
\emph{Schneider Electric Taiwan Co., Ltd}\\

\noindent
This report describes a 6-model ensemble with multi-scale test-time augmentation (TTA), low-threshold pseudo-labeling, semi-supervised learning via MixPL~\cite{chen2023mixpl}, and CLAHE-based TTA diversity, fused via Weighted Boxes Fusion (WBF)~\cite{solovyev2021wbf}.

\textbf{Models.}
Five distinct architectures are combined for prediction diversity: Co-DINO~\cite{zong2023codino} (Swin-L, O365$\rightarrow$COCO pretrained, val~AP~$\approx$0.456), DDQDETR~\cite{zhang2023ddqdetr} (Swin-L, val~0.434), DINO~\cite{zhang2023dino} (Swin-L, val~0.428), RTMDet-l~\cite{lyu2022rtmdet} (CSPNeXt, val~0.437), and RF-DETR~\cite{robinson2026rfdetr} (DINOv2 ViT, val~0.442). A 6th model---Co-DINO trained with MixPL teacher-student EMA (val~0.472)---provides the semi-supervised component. All models except RF-DETR use MMDetection~3.3.0.

\textbf{Training.}
Training proceeds in three phases on a single A100~40GB GPU. Phase~1 ($\sim$10\,h): five base models are trained on the 704 labeled images with AdamW (lr=$10^{-4}$, wd=$5 \times 10^{-4}$), cosine schedule, EMA, and RandomResize (0.5--1.5$\times$) + flip augmentation at resolutions $1333 \times 1000$ (DETR variants) and $960 \times 720$ (RTMDet). Phase~2 ($\sim$5\,h): pseudo labels are generated on the 381 test images from ensemble predictions at per-class thresholds (vessel~$\geq$0.35, nav~$\geq$0.25), yielding $\sim$794 pseudo annotations; all five models are fine-tuned on 1085 images (704 GT + 381 pseudo). Phase~3 ($\sim$2\,h): MixPL teacher-student EMA (momentum=0.0002) is applied to Co-DINO for 4000 iterations (batch size~2, gradient accumulation~4), producing the 6th model with val~AP~0.472 (+0.016 over Phase~2).

\textbf{Inference \& Ensemble.}
Each model runs TTA at 4 scales $\times$ 2 orientations = 8 sources. For the two Co-DINO models, CLAHE preprocessing (clipLimit=3.0, tile=$8 \times 8$) adds 16 extra sources. The resulting 64 sources (48 standard + 16 CLAHE) are fused via WBF (equal weights, iou\_thr=0.74, skip=0.15). Full inference takes $\sim$2--3 min/image on an A100.

\textbf{Domain observations.}
63.4\% of objects are $<32 \times 32$~px (nav median $8.3 \times 18.6$~px). Higher inference resolution is the single largest lever. Architecture diversity (DETR + anchor-based + ViT) provides complementary predictions. CLAHE improves strong models but hurts weaker ones ($-0.002$ to $-0.008$~AP). Semi-supervised MixPL gave the largest late-stage gain (+0.0033~AP). Only the competition dataset was used; pretraining utilized COCO, Objects365 (Co-DINO), and ImageNet-22K (Swin-L).

\subsection{\texorpdfstring{\protect\msecond{}}{(2nd)} Optimization-Diverse Multi-Scale Ensemble}
\label{thermal-det:deimv2-ensemble}
\noindent
\emph{Chun-Ming Tsai$^1$, Jun-Wei Hsieh$^2$, Ming-Ching Chang$^3$}\\
\texttt{cmtsai@go.utaipei.edu.tw, jwhsieh@nycu.edu.tw, mchang2@albany.edu}\\
\emph{$^1$University of Taipei, $^2$National Yang Ming Chiao Tung University, $^3$University at Albany, SUNY}\\

\noindent
This report presents an optimization-diverse multi-scale ensemble framework for thermal object detection. Multiple DEIMv2 detectors are trained with different random seeds and input resolutions, and their predictions are combined via WBF~\cite{solovyev2021wbf} to improve robustness and localization accuracy.

\textbf{Architecture.}
All models are based on the DEIMv2 detection framework with a DINOv3-X transformer backbone, implemented in PyTorch. The final system ensembles 11 detectors operating at four resolutions ($1024 \times 1024$, $1280 \times 1280$, $1440 \times 1440$, and $1600 \times 1600$). Predictions from all models are merged using WBF with parameters IoU=0.72, skip\_box\_thr=0.004, and per-class top-$k$=250. Higher-resolution models are assigned slightly larger fusion weights.

\textbf{Training.}
All models are initialized from the publicly available DEIMv2-X COCO pretrained checkpoint, which uses a DINOv3 backbone pretrained in a self-supervised manner on large-scale datasets (e.g., LVD). No additional external datasets beyond these publicly available pretrained weights are used; all fine-tuning is performed solely on the MaCVi 2026 thermal dataset. Training employs multi-scale augmentation, random horizontal flipping, geometric transformations, and photometric distortions. CLAHE-based contrast enhancement is applied only in the 1440-resolution configuration as a probabilistic augmentation ($p = 0.5$) during training; it is not used during inference.

\textbf{Inference.}
Experiments are conducted on a workstation with an NVIDIA RTX 3090. Single-model inference takes approximately 0.2--0.3\,s per image; the 11-model ensemble adds additional overhead due to multi-model inference and fusion.

\subsection{\texorpdfstring{\protect\mthird{}}{(3rd)} AGAF (Agreement-Gated Auxiliary Fusion)}
\label{thermal-det:agaf}
\noindent
\emph{Wonwoo Jo, Hyewon Chun, Sangmun Lee}\\
\texttt{\{wonwoojo, hyewonchun, sangmunlee\}@hd.com}\\
\emph{HD Korea Shipbuilding \& Offshore Engineering Co., Ltd.}\\

\noindent
This report presents AGAF, a pipeline combining RF-DETR~\cite{robinson2026rfdetr} 2XLarge detectors with annotation refinement, skyline filtering, and agreement-gated auxiliary fusion via class-aware WBF~\cite{solovyev2021wbf}.

\textbf{Architecture.}
The system is built on RF-DETR 2XLarge with a DINOv2-Base~\cite{oquabdinov2} backbone and windowed attention. Multiple checkpoints trained at 960\,px and 1280\,px resolutions are combined into a 9-source ensemble via class-aware WBF (IoU\,=\,0.76) with per-class weights for vessels and navigational objects, reflecting the different transfer characteristics of each category.

\textbf{Domain-specific adaptations.}
Four key adaptations target the thermal maritime domain: (1)~\emph{Annotation refinement}: the provided labels contain inconsistent wind-turbine annotations; correcting these and producing \emph{turbine-added} (primary) and \emph{turbine-removed} (auxiliary) annotation variants yields the single largest gain (+0.027~AP). (2)~\emph{Skyline filtering}: a YOLO-based~\cite{Jocher_Ultralytics_YOLO_2023} horizon detector, trained on manually annotated skyline boxes from the challenge images, suppresses navigational-object false positives above the estimated horizon (+0.012~AP). (3)~\emph{Class-aware ensemble}: per-class WBF source weights account for category-dependent detection difficulty. (4)~\emph{Agreement-gated fusion}: an auxiliary model trained on the turbine-removed annotations confirms class-1 (vessel) hypotheses asymmetrically---it can only confirm, not create, detections, avoiding label-policy mismatch between annotation variants.

\textbf{Training.}
All models use AdamW with base LR $10^{-4}$ and encoder LR $1.5 \times 10^{-4}$, step LR decay at epoch 40, ViT layer decay 0.8, EMA (decay 0.993), and multi-scale training with expanded scales. Training runs range from 10 to 50 epochs depending on the variant. Only the MaCVi 2026 dataset is used (667 train / 162 val images); no external maritime data or annotations are employed. The DINOv2 backbone uses standard ImageNet-pretrained initialization.

\textbf{Inference.}
Experiments are conducted on a single NVIDIA RTX 5070 (12\,GB). Single-model inference runs at $\sim$15\,FPS at 960\,px; the full 9-source ensemble pipeline operates at $\sim$1.5\,FPS per image.

\section{LaRS Panoptic Segmentation Challenge}
\label{panseg:appendix}
\subsection{\texorpdfstring{\protect\mfirst{}}{(1st)} M2F-DINOv3}
\label{panseg:m2f-dino}
\noindent
\emph{Ivan Martinović}\\
\texttt{ivan.martinovic@fer.hr}\\
\emph{Faculty of Electrical Engineering and Computing, University of Zagreb, Croatia}\\
Our method is based on Mask2Former~\cite{cheng2021mask2former} with a pre-trained DINOv3~\cite{simeoni2025dinov3} vision transformer backbone. The Mask2Former segmentation head expects a multi-scale feature pyramid, whereas the DINOv3 backbone produces features at a native stride of $16$, denoted by $\mathbf{F}_{16}$. To bridge this mismatch, we construct a feature pyramid following the design used in ViTDet~\cite{li2022exploring} and EoMT~\cite{kerssies2025your}.
More specifically, we build feature maps at strides $4$, $8$, $16$, and $32$, corresponding to $\mathbf{F}_{4}$, $\mathbf{F}_{8}$, $\mathbf{F}_{16}$, and $\mathbf{F}_{32}$. Starting from $\mathbf{F}_{16}$, we obtain higher-resolution features using repeated upsampling blocks, and lower-resolution features using downsampling blocks. Each block consists of a $2 \times 2$ convolution (transposed for upsampling, standard for downsampling) with stride $2$, followed by a GELU activation, a depthwise $3 \times 3$ convolution, and a normalization layer. In our implementation, the same stride-$16$ DINOv3 feature tensor is reused to generate all pyramid levels. The pixel decoder and transformer decoder follow the original Mask2Former design.
We evaluate two backbone variants, DINOv3-L and DINOv3-H+. The models are trained on four NVIDIA RTX A6000 Ada GPUs (48\,GB VRAM per GPU) with a total batch size of $16$. Training is performed for $90{,}000$ iterations using a crop size of $512 \times 1024$. Unless otherwise stated, the remaining optimization and decoder hyperparameters follow the standard Mask2Former configuration for Cityscapes.
To improve performance on underrepresented categories, we apply rare-class oversampling. Following the repeat-factor sampling strategy introduced for LVIS~\cite{gupta2019lvis}, we first compute the image-level frequency $f(c)$ of each class $c$, i.e., the fraction of training images in which class $c$ appears. We then define the class repeat factor as
\[
r(c) = \max \left( 1, \sqrt{\frac{t}{f(c)}} \right),
\]
where $t$ is the repeat threshold. The repeat factor assigned to an image is the maximum repeat factor among all classes present in that image. In our experiments, we use $t = 0.25$.
In addition, we use a rare-class-aware cropping strategy. For images containing rare classes, we first sample a target segment with probability proportional to its class repeat factor. We then draw up to $10$ random crops and accept the first one that contains at least $10\%$ of the target segment area. If none of the sampled crops satisfies this condition, we keep the last sampled crop. This procedure makes rare classes more likely to appear in the final training crop.
\begin{table}[t]
\centering
\caption{Panoptic segmentation results on the LARS validation set.}
\label{tab:lars_val_results}
\begin{tabular}{lccc}
\hline
Method & PQ & SQ & RQ \\
\hline
Mask2Former + DINOv3-L  & 54.8 & 75.8 & 63.5 \\
Mask2Former + DINOv3-H+ & 55.8 & 75.5 & 65.4 \\
\hline
\end{tabular}
\end{table}
Table~\ref{tab:lars_val_results} reports validation-set results for the two backbone variants. For the final submission, we retrained the Mask2Former + DINOv3-H+ model on the union of the training and validation splits. The resulting model achieved $53.5$ PQ on the LARS test set.

\subsection{\texorpdfstring{\protect\msecond{}}{(2nd)} MaskDINOv3}
\label{panseg:maskdino}
\emph{Jannik Sheikh\textsuperscript{1}, Andreas Michel\textsuperscript{1}, 
   Wolfgang Gross\textsuperscript{1}, 
   Martin Weinmann\textsuperscript{2}}\\ 
\texttt{firstname.lastname@iosb.fraunhofer.de} \\
\texttt{martin.weinmann@kit.edu} \\
\emph{\textsuperscript{1}Fraunhofer IOSB, Germany \\ 
\textsuperscript{2}Karlsruhe Institute of Technology, Germany} \\
This technical report describes our solution to the $4^{\text{th}}$ Workshop on Maritime Computer Vision (MaCVi) USV-based Panoptic Segmentation challenge, building upon our previous top-3 submission (PQ 53.9) using MaskDINO~\cite{Li2022Mask}.

This work evaluates DINOv3 \cite{simeoni2025dinov3}, a vision foundation model for dense features, as a replacement for the ImageNet-22K~\cite{ridnik2021imagenet} pre-trained Swin-L \cite{liu2021swin} backbone. Since MaskDINO requires multi-scale features but DINOv3 produces single-scale outputs, we employ DEIMv2 \cite{huang2025real}, which uses a Spatial Tuning Adapter (STA) to convert DINOv3 outputs into multi-scale features.

\textbf{Experimental Settings.}
Experiments were conducted on four NVIDIA H100 GPUs using a distilled DINOv3 ViT-H+/16 model (840M parameters). Features were extracted from ViT blocks 14, 22, and 31 and passed to the STA module, which used a base channel dimension of 32 for its convolutions. Training with AdamW \cite{loshchilov2018decoupled} followed two stages: (1) frozen backbone, training STA and MaskDINO head for 25k iterations with base LR $2.2 \times 10^{-5}$; (2) full fine-tuning in two runs with progressively reduced LR ($2.2 \times 10^{-5}$ then $5 \times 10^{-6}$) and backbone multipliers (0.1, 0.05) for 15k and 10k iterations respectively, along with per-class weighting to improve PQ\textsuperscript{Th}. Both stages used augmentations including $1024\times1024$ crops, horizontal flips, and multi-scale resizing.

\textbf{Observations and Remarks.}
On LaRS~\cite{Zust2023LaRS} test, our model achieves PQ 48.3 (F1 69.5). Compared to MaskDINO with Swin-L, PQ\textsuperscript{St} improved notably (92.3 $\rightarrow$ 94.9), indicating DINOv3's suitability for stuff segmentation. However, PQ\textsuperscript{Th} declined (39.4 $\rightarrow$ 30.9) despite per-class weighting. We hypothesize this stems from pre-training differences: Swin-L benefits from COCO panoptic pre-training providing instance-level priors, while DINOv3's self-supervised training may lack instance-discriminative representations. The PQ\textsuperscript{St} gain highlights DINOv3's potential, warranting further research on thing-class adaptations.

\subsection{\texorpdfstring{\protect\mthird{}}{(3rd)} ThingSeg-DGCR}
\label{panseg:thingseg}
\emph{Hyewon Chun, Wonwoo Jo, Sangmun Lee}\\
\texttt{\{hyewonchun, wonwoojo, sangmunlee\}@hd.com}\\
\emph{HD Korea Shipbuilding \& Offshore Engineering Co., Ltd.}\\
\textbf{Algorithm outline}. We use two MaskDINO-R50 checkpoints
as the panoptic backbone and an RF-DETR-Seg
Medium branch for thing recovery, building on the official
MaskDINO~\cite{Li2022Mask} and RF-DETR~\cite{robinson2026rfdetr} codebases. One
MaskDINO checkpoint is trained with rare-class rebalancing
and the other with thing-aware crop fine-tuning; both are
evaluated at scale 896 with horizontal flips. The RF-DETR
branch runs only on the eight thing classes at resolution
720. We merge detector masks into the panoptic output only
when conservative overlap rules are satisfied, then apply
class-specific GrabCut refinement.

\textbf{Training}. The MaskDINO models start from Detectron2
ImageNet-pretrained ResNet-50 weights and use batch size
1 with large-scale jitter at 896. We then run 16k iterations
with rare-class repeat sampling and 8k more with thingaware
crops. RF-DETR-Seg Medium uses EMA, gradient
checkpointing, batch size 2 with gradient accumulation 8,
and offline copy-paste of rare thing instances onto water
regions, initialized from an earlier LaRS RF-DETR model.

\textbf{Datasets}. We used only official LaRS data and derivatives:
the panoptic train/validation/test splits, a thing-only COCO
export derived from LaRS, and synthetic copy-paste images
built from LaRS training images. We did not use external
maritime datasets, private annotations, or external imagery.
The ResNet-50 backbone uses standard ImageNet pretraining.

\textbf{Hardware and inference speed}. All training and inference
were run on a single NVIDIA RTX A4000 GPU. A
single MaskDINO checkpoint runs at about 3.5–4.5 FPS; the
full pipeline runs at roughly 0.5–0.7 FPS.

\textbf{Domain-specific adaptations}. In LaRS, stuff is relatively
easy; the main challenge is recovering small maritime thing
instances. Buoys, swimmers, paddle boards, row boats, and
rare obstacles are sparse, small, and visually ambiguous, so
we treated the task primarily as a thing-recovery problem.
That led us to rare-class rebalancing, thing-aware crops, a
conservative detector branch, and stronger GrabCut settings
for buoys and swimmers.

\section{LaRS Embedded Segmentation Challenge}
\subsection{\texorpdfstring{\protect\mfirst{}}{(1st)} DSOS-Net: DINOv3-based Water Surface Obstacle Segmentation Network}
\label{embseg:dsos-net}
\noindent
\emph{Yuan Feng}\\
\texttt{fengyuan9822@outlook.com, 931772830@qq.com}\\
\emph{Independent Researcher}\\

\textbf{Method.} Inspired by the encoder-decoder architecture, we propose DSOS-Net for real-time surface obstacle detection. The network architecture primarily consists of an encoder section and a decoder section. In the encoder, we employ ConvNeXt \cite{liu2022convnet} as the backbone network, which has been self-supervised pre-trained on the LVD-1689M dataset using the DINOv3 method \cite{simeoni2025dinov3}, enabling effective extraction of robust multi-scale features. The decoder adopts the RSOS-Net \cite{wang2026rsosnet} architecture, which comprises a lightweight feature pyramid network, a fast pyramid pooling module, and an attention-based feature fusion module. Specifically, the fast pyramid pooling module effectively expands the receptive field by combining global average pooling and cascaded average pooling operations, enabling the capture of both global and local contextual information, which is crucial for distinguishing water surface disturbances from real obstacles in maritime environments. The attention-based feature fusion module, employing a channel-spatial attention mechanism, enables DSOS-Net to focus more on real obstacles, reducing false positives and missed detections. The combination of the robust feature representation from ConvNeXt and the efficient segmentation head from RSOS-Net achieves a balance between accuracy and inference speed.

\textbf{Training.} We implemented DSOS-Net using PyTorch with an image input size of $768 \times 384$ and a batch size of 6. The training was conducted on an NVIDIA RTX 2080 Ti GPU with 12 GB memory. We adopted a two-stage training strategy. In the first stage, the model was trained for 100 epochs using the AdamW optimizer with an initial learning rate of $1 \times 10^{-4}$ and a cosine learning rate scheduler. The loss function combines cross-entropy loss and Dice loss with equal weights. To address the severe class imbalance problem, we set the class weights for obstacle, water, and sky classes as 3.0, 1.0, and 1.0, respectively. In the second stage, we continued training for an additional 20 epochs, resulting in a total of 120 epochs, with equal class weights of 1.0 for all classes to achieve better precision-recall balance. For data augmentation, techniques including random horizontal flip with probability 0.5, random vertical flip with probability 0.3, random rotation within $\pm 15^{\circ}$, color jitter with brightness, contrast, and saturation adjustments within $\pm 0.3$ and hue within $\pm 0.15$, and Gaussian blur were applied. Regarding datasets, only the LaRS dataset \cite{Zust2023LaRS} was utilized for training and validation. The backbone network was initialized with ConvNeXt weights pretrained by the DINOv3 method.

\textbf{Observations}
\begin{itemize}
    \item \textbf{Two-stage Training Effect:} In our submissions, the version with ID 16965 from the first stage achieved a Q-score of 61.5, F1-score of 65.7, and mIoU of 93.5, with precision of 56.7\% and recall of 78.2\% at 55.9 FPS. The high class weight for the obstacle class resulted in high recall but relatively low precision. The version with ID 17010 from the second stage achieved the best Q-score of 61.9, F1-score of 66.0, and mIoU of 93.8, with precision of 62.6\% and recall of 69.7\% at 66.5 FPS, ranking 1st on the leaderboard. The inference speed was evaluated by submitting ONNX models to the official evaluation server.

    \item \textbf{Backbone Efficiency:} With a smaller ConvNeXt variant, DSOS-Net can achieve up to 98 FPS while maintaining competitive segmentation accuracy.

    \item \textbf{Feature Representation:} The self-supervised pre-training of DINOv3 on large-scale datasets provides robust feature representations that transfer well to maritime obstacle segmentation, effectively reducing the domain gap between pre-training and downstream tasks.

    \item \textbf{Evaluation Strictness:} The evaluation of the USV-based Embedded Obstacle Segmentation challenge appears to be stricter compared to last year.

    \item \textbf{Code Availability:} The code for DSOS-Net will be publicly accessible soon at \href{https://github.com/Yuan-Feng1998}{https://github.com/Yuan-Feng1998}.
\end{itemize}

\subsection{\texorpdfstring{\protect\msecond{}}{(2nd)} PIDNet-S with Copy-Paste Obstacle Augmentation}
\label{embseg:pidnet-s}
\noindent
\emph{Jose Mateus Raitz Persch\textsuperscript{1}, Rahul Harsha Cheppally\textsuperscript{2}}\\
\texttt{jmraitzp@protonmail.com, r4hul@ksu.edu}\\
\emph{\textsuperscript{1}Independent Researcher, \textsuperscript{2}Kansas State University}\\

\textbf{Architecture.}
We use PIDNet-S~\cite{xu2023pidnet} (\href{https://github.com/XuJiacong/PIDNet}{github.com/XuJiacong/PIDNet}), a lightweight three-branch (Proportional--Integral--Derivative) network with 32 base channels, PPM pooling (96~channels), and a PIDHead decoder (128~channels, 3~classes). The Derivative branch explicitly models boundary detail, which is particularly relevant for maritime segmentation where water--obstacle boundaries are often ambiguous. The backbone is initialized from ImageNet-1K pretrained weights. We made no architectural modifications---PIDNet-S was selected specifically because its BatchNorm + Conv + ReLU composition quantizes cleanly under the server's INT8 quantization pipeline on the Luxonis RVC4. During our evaluation phase, we found that architectures relying on attention mechanisms (SegFormer) or large-kernel depthwise convolutions (SegNeXt) suffered catastrophic INT8 degradation ($-$67\% to $-$82\% mIoU), making simple convolution-based designs the only viable choice.

\textbf{Training.}
We train at the target resolution of $768 \times 384$ using SGD (lr\,=\,0.133, momentum\,=\,0.9, weight decay\,=\,$5{\times}10^{-4}$) for 15{,}000 iterations with polynomial LR decay (power\,=\,0.9) and a 2{,}500-iteration linear warmup.
Batch size is 320 (160 per GPU). Losses: CrossEntropy ($w$\,=\,0.4), two OHEM losses ($w$\,=\,1.0, threshold\,=\,0.9), and BoundaryLoss ($w$\,=\,20). Augmentations include Large Scale Jittering (ratio $\in [0.5, 2.0]$), random crop, horizontal flip, enhanced photometric distortion (brightness, contrast, saturation, hue, gamma jitter), and reflect padding matching the server preprocessing. Framework: mmsegmentation~\cite{mmseg2020}, PyTorch~2.1.2, CUDA~12.1.

\textbf{Copy-Paste augmentation.}
The key component of our method is a Copy-Paste obstacle augmentation~\cite{ghiasi2021copypaste} using crops extracted from three external maritime datasets: \textbf{WaterScenes}~\cite{yao2024waterscenes}, \textbf{ROSEBUD}~\cite{lambert2022rosebud}, and the night subset of \textbf{MULTIAQUA}~\cite{iancu2024multiaqua}. All three were remapped to the LaRS 3-class schema and deduplicated with perceptual hashing. We mined ${\sim}$30{,}000 obstacle crops (water-adjacent, area 50--100k\,px at target resolution) and stored them in a shared database. During training, with probability 0.7, 1--3 crops are sampled with size-based tier weights (oversampling small obstacles), brightness-matched to the local water region, and pasted before photometric distortion. This approach was motivated by a failure analysis of our Phase~2 baseline server submission and a spatial out-of-distribution study on the LaRS test set, both of which showed that the majority of false negatives involved small or rare obstacle types under unusual lighting---gaps that external maritime datasets could address. An initial run with a shorter schedule (8k iterations) failed due to ``augmentation shock,'' which we resolved by extending to 15k iterations with longer warmup.

\textbf{Datasets.}
LaRS~\cite{Zust2023LaRS} (train: 2{,}605 images) for training; ImageNet-1K for backbone pretraining; WaterScenes, ROSEBUD, and MULTIAQUA\_night as external augmentation sources (crops only, no joint training). No custom-annotated data was used.

\textbf{Hardware \& inference speed.}
Training was performed on 2$\times$ NVIDIA RTX A6000 (48\,GB) with DDP. On the challenge server (Luxonis RVC4, INT8 quantization), our model achieves 67.7\,FPS.

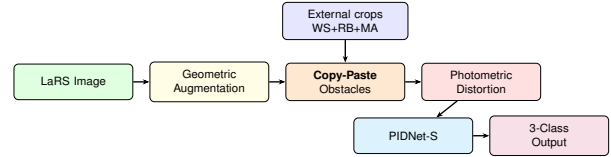
\begin{figure}[t]
\centering
\resizebox{0.95\columnwidth}{!}{%
\begin{tikzpicture}[
  box/.style={draw, rounded corners=2pt, minimum height=0.55cm,
              text width=1.6cm, align=center, font=\tiny\sffamily},
  arr/.style={-{Stealth[length=3pt]}, semithick},
  dbbox/.style={draw, rounded corners=2pt, minimum height=0.45cm,
                fill=blue!10, text width=1.7cm, align=center, font=\tiny\sffamily},
]
  \node[box, fill=green!12] (input) {LaRS Image};
  \node[box, fill=yellow!12, right=0.25cm of input] (geom) {Geometric\\Augmentation};
  \node[box, fill=orange!18, right=0.25cm of geom] (cp) {\textbf{Copy-Paste}\\Obstacles};
  \node[box, fill=red!12, right=0.25cm of cp] (photo) {Photometric\\Distortion};
  \node[box, fill=cyan!12, below=0.55cm of $(cp)!0.5!(photo)$] (model) {PIDNet-S};
  \node[box, fill=purple!12, right=0.25cm of model] (out) {3-Class\\Output};
  \node[dbbox, above=0.35cm of cp] (db) {External crops\\WS+RB+MA};
  \draw[arr] (input) -- (geom);
  \draw[arr] (geom) -- (cp);
  \draw[arr] (cp) -- (photo);
  \draw[arr] (db) -- (cp);
  \draw[arr] (photo) -- (model);
  \draw[arr] (model) -- (out);
\end{tikzpicture}%
}
\caption{Training pipeline. WS\,=\,WaterScenes, RB\,=\,ROSEBUD, MA\,=\,MULTIAQUA\_night.}
\label{fig:pipeline}
\end{figure}

\subsection{\texorpdfstring{\protect\mthird{}}{(3rd)} RSOS-Net R50 PyTorch + YOLOX-HR}
\label{embseg:rsos-net}
\noindent
\emph{Justin Davis, Mehmet E. Belviranli}\\
\texttt{\{jcdavis, belviranli\}@mines.edu}\\
\emph{Colorado School of Mines, Colorado, United States}\\

\textbf{Method.}
Our submission is a PyTorch reimplementation of RSOS-Net~\cite{wang2026rsosnet}, the 1st-place solution from the MaCVi~2025 Embedded Obstacle Segmentation Challenge~\cite{Kiefer_2025_WACV}.
RRSOS-Net is a lightweight encoder-decoder segmentation network designed for water-surface obstacle detection, using multi-scale context pooling and channel-spatial attention to suppress false positives from reflections, sun glitter, and wakes.  
In the MaCVi~2025 challenge proceedings, the authors submitted both a ResNet-50 variant (Q\,=\,63.6, 102.8~FPS) and a ResNet-101 variant (Q\,=\,64.2, 85.1~FPS), with the R101 achieving the top accuracy score.
The subsequent journal paper~\cite{wang2026rsosnet} provides a comprehensive description of the architecture using a ResNet-18 backbone.
Our reimplementation is based on this full architectural specification, adapting the design to a ResNet-50 backbone with a 4-scale feature pyramid (layers~1--4) and applying a revised training pipeline.
The ResNet-50 backbone is initialized with ImageNet-1K pretrained weights and uses output stride~16 via dilated convolutions in stage~4.

\textbf{Training.}
All training used only the LaRS~\cite{Zust2023LaRS} dataset (SGD, 16-bit mixed precision, batch size~8, 200~epochs).
The loss combines cross-entropy, dice loss (weight~2.0), and focal loss (weight~1.0) on the main head, with standard cross-entropy on two auxiliary FCN heads (weight~0.4).
Augmentations include random scale/crop/flip, color jitter, brightness--contrast and HSV shifts, Gaussian and motion blur, noise, random shadows, and coarse dropout.
Training was performed on a single NVIDIA RTX~5080 GPU.
The original authors trained for 160,000 steps at batch size~16 (${\sim}1{,}000$ epochs), roughly $5\times$ our training volume, which likely accounts for some of the accuracy gap between our R50 (Q\,=\,46.4) and theirs (Q\,=\,63.6).

\textbf{Additional Submission---YOLOX-HR.}
We also submitted YOLOX-HR~\cite{yolox2021}, a custom high-resolution variant of YOLOX using the medium CSPDarknet53 backbone (YXHR\_medium), motivated by the observation that small dynamic obstacle detection was critical for F1/Q~metrics.
YOLOX-HR employs a stacked dual Path Aggregation FPN (PAFPN) that fuses features from stride~2 through~32, yielding predictions at stride~2 to preserve spatial detail for small objects.
Its purely convolutional architecture (Conv-BN-SiLU) typically has highly optimized kernels on embedded platforms such as the RVC4.

\textbf{Observations.}
\begin{itemize}
    \item RSOS-R50 achieves Q\,=\,46.4 at 96.3~FPS, while YXHR\_medium achieves Q\,=\,45.3 at 64.9~FPS. Despite YOLOX-HR having more parameters and higher mIoU, the overall Q is lower at a significant FPS cost, showing that RSOS-Net's lightweight decoder modules scale more efficiently at nearly $1.5\times$ the inference speed.
    \item YOLOX-HR showed better Q~scores than our RSOS-Net variant during training but overfit to the train/val split, most likely due to the low amount of training data relative to the parameter count.
    \item The accuracy gap between our R50 and the original authors' R50 is likely attributable to: training duration ($5\times$ fewer samples), adaptation of the published R18 architecture to an R50 backbone, and training pipeline differences.
\end{itemize}

\section{Multimodal Semantic Segmentation Challenge}

\label{mmseg:appendix}

\subsection{\texorpdfstring{\protect\mfirst{}}{(1st)} GatedMemorySAM}
\label{mmseg:GatedMemorySAM}
\emph{Jemo Maeng, Sangmin Park, Seongju Lee, Kyoobin Lee}\\
\texttt{\{maengjemo, leowiu24, lsj2121\}@gm.gist.ac.kr, kyoobinlee@gist.ac.kr}\\
\emph{GIST AI LAB, Gwangju Institute of Science and Technology, South Korea}

\paragraph{Method.}
We introduce \textbf{GatedMemorySAM}, which builds upon MemorySAM~\cite{liao2025memorysam}
that repurposes SAM2's~\cite{ravi2024sam2} temporal memory attention for cross-modal fusion by
treating each input modality (RGB, LiDAR, Thermal) as a separate ``frame'' in SAM2's video
pipeline. We extend this framework with two key modifications:
(1) Soft MoE LoRA adaptation, and
(2) quality-aware modality scoring.

For backbone adaptation, we replace standard LoRA~\cite{hu2022lora} with Soft MoE routing~\cite{puigcerver2024softmoe}
over multiple LoRA experts, inserted into every attention block (Q and V projections, 48 layers total)
with rank $= 4$ and 3 experts. Each spatial token is softly routed to all experts via a learned gating
network, enabling per-token specialization across modalities.

\begin{figure}[t]
    \centering
    \includegraphics[width=\linewidth]{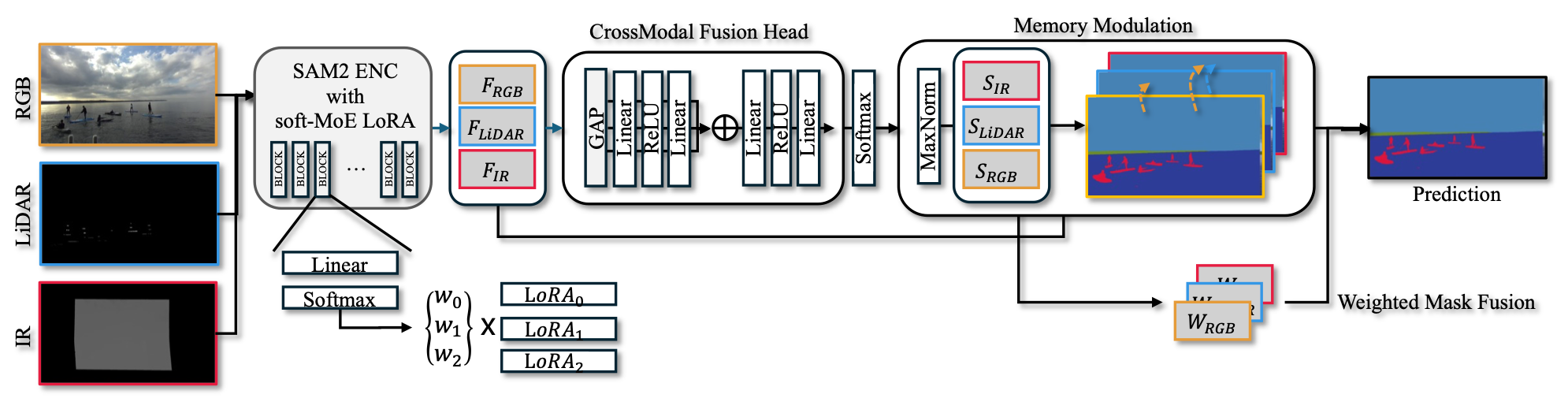}
    \caption{Overview of GatedMemorySAM. Each modality is encoded by SAM2 with Soft-MoE LoRA,
    scored by the CrossModalFusionHead, and fused via memory modulation and weighted mask fusion.}
    \label{fig:overview}
\end{figure}

For modality weighting, we introduce a \texttt{CrossModalFusionHead} that compares all modality features
via global average pooling and a shared comparison MLP to produce per-modality softmax weights.
These weights drive two score-based fusion mechanisms:
(i) \emph{memory modulation}, which max-normalizes the scores and scales each modality's backbone features
before they enter SAM2's memory bank, so that the best modality retains full strength while weaker ones
are suppressed; and
(ii) \emph{weighted mask fusion}, which averages the per-modality decoder outputs using the same softmax weights.
The overall pipeline is illustrated in Figure~\ref{fig:overview}.

\paragraph{Training Details.}
We train with the AdamW optimizer (lr $= 6 \times 10^{-4}$, weight decay $= 0.01$) using a warmup
polynomial LR schedule (10-epoch warmup, power $= 0.9$) for 200 epochs.
The loss function is Online Hard Example Mining Cross-Entropy (OHEM-CE).
Input images are resized to $1024 \times 1024$.
We use DDP training with an effective batch size of 16 across 8 GPUs with gradient accumulation.

\paragraph{Night Augmentation.}
Since the test set consists entirely of nighttime images while training data is daytime, we apply an
extensive nighttime simulation pipeline: brightness darkening (range $[0.03, 0.45]$ with 60\% dark-biased sampling),
contrast reduction, gamma correction ($[0.4, 0.8]$), Gaussian and Poisson shot noise, cross-modal replacement
(CRM, $p = 0.20$), and PhysAug~\cite{xu2025physaug}-inspired spatial perturbations
(random convolution filters and planar Fourier wave patterns, $p = 0.40$).

\paragraph{Datasets and Pretrained Weights.}
We train exclusively on the MULTIAQUA dataset provided by the challenge organizers.
The SAM2 Hiera-B+ backbone is initialized from publicly available pretrained weights.
No additional external datasets or annotations are used.

\paragraph{Hardware and FPS.}
Training was conducted on $8 \times$ NVIDIA RTX 3090 GPUs (24GB each).
Inference runs on a single NVIDIA Titan RTX GPU at approximately 2.1 FPS
($1024 \times 1024$ input, 3 modalities processed sequentially).

\paragraph{Maritime Domain Adaptations.}
The primary challenge is the extreme day-night domain gap (daytime training vs.\ nighttime-only test).
Our nighttime simulation augmentation was the single most impactful component,
nearly doubling the test mIoU.


\subsection{\texorpdfstring{\protect\msecond{}}{(2nd)} Adapted MFNet-SAM-LoRA}
\label{mmseg:MFNet}

\emph{Yusi Cao, Jiahui Wang, Lingling Li, Xu Liu, LiCheng Jiao}\\
\texttt{cys734511@163.com, 18257864149@163.com, llli@xidian.edu.cn, xuliu361@163.com, lchjiao@mail.xidian.edu.cn}\\
\emph{School of Artificial Intelligence, Xidian University, China}

\textbf{Method.} We adapted the existing Multimodal Fine-tuning Network (MFNet) by leveraging the Segment Anything Model (SAM, ViT-Large) as our foundational backbone. To process three distinct modalities within MFNet's dual-branch architecture, we feed the RGB image into the primary branch, while concatenating the Thermal image and projected LiDAR data (distance d and reflectivity r) into a 3-channel auxiliary tensor for the secondary branch. Both inputs are processed concurrently by the frozen SAM encoder, which is equipped with trainable Low-Rank Adaptation (LoRA) layers for parameter-efficient tuning. The extracted features are then fused across scales via Squeeze-and-Excitation (SE) blocks and decoded using a UNet-style head. Observing the drastic day-to-night domain shift in the maritime environment, this Thermal-LiDAR fusion ensures robustness against nighttime RGB degradation. Furthermore, to handle recording boat artifacts, we explicitly set \texttt{ignore\_index=0} in our loss computation to prevent erroneous structure learning.

\textbf{Training.} Our training strictly utilized the daytime splits of the provided MULTIAQUA dataset, relying solely on SAM's native SA-1B pre-trained weights without any additional maritime datasets or custom pseudo-labels. We optimized the network using AdamW (initial LR=0.001, weight decay=1e-4) with a CosineAnnealingWarmRestarts scheduler over 80 epochs. Data augmentation was minimal, utilizing only 512×512 random cropping and horizontal flipping. To manage hardware constraints, Automatic Mixed Precision (AMP) and gradient accumulation were employed to achieve an effective batch size of 8. All procedures were executed on a single NVIDIA GeForce RTX 4090 GPU, achieving an efficient inference speed of approximately 12 FPS.

\subsection{\texorpdfstring{\protect\mthird{}}{(3rd)} Modified DustNet}
\label{mmseg:DustNet}

\emph{Andreas Michel\textsuperscript{1}, Jannik Sheikh\textsuperscript{1}, Jannick Kuester\textsuperscript{1}, Bettina Felten\textsuperscript{1}, Wolfgang Gross\textsuperscript{1}, Martin Weinmann\textsuperscript{2}}\\ 
\texttt{firstname.lastname@iosb.fraunhofer.de} \\
\texttt{martin.weinmann@kit.edu} \\
\emph{\textsuperscript{1}Fraunhofer IOSB, Germany \\ 
\textsuperscript{2}Karlsruhe Institute of Technology, Germany} \\

\textbf{Methodology.} 
Our methodology builds upon the DustNet architecture \cite{michel2023dustnet,michel2025dustnet++}, a deep neural network specifically designed for visual density estimation tasks. The DustNet-C variant leverages a dual-branch encoder architecture to extract temporal, global, and local feature representations.

We adapt this architecture for multi-modal sensor fusion by repurposing the dual input streams. Hereby, the primary branch processes RGB imagery, while the secondary branch receives a fused representation of LiDAR and thermal infrared data. For the latter, 2D projected LiDAR data is concatenated channel-wise with the thermal infrared image, preprocessed using Contrast Limited Adaptive Histogram Equalization (CLAHE) \cite{zuiderveld1994contrast}. Each modality stream is processed through dedicated backbone networks for domain-specific feature extraction. The resulting feature maps are subsequently integrated via a cross-attention mechanism.

\textbf{Experimental Settings.}
All experiments were conducted using two Nvidia A100 GPUs, each equipped with 80 GB of VRAM. The modified DustNet architecture was implemented with a Swin-L \cite{liu2021swin} backbone, initialized with weights pretrained on the ImageNet-1K dataset \cite{deng2009imagenet} to leverage learned visual representations. The model was trained on the maritime subset of the MULTIAQUA dataset for 36 epochs, employing cross-entropy as the loss function and AdamW \cite{loshchilov2017decoupled} as the optimizer. Data augmentation during this phase was limited to large-scale jitter and stochastic horizontal flipping. Subsequently, fine-tuning was performed for 2 additional epochs, incorporating stochastic brightness attenuation on the RGB input channels to enhance robustness under variable illumination conditions. The trained model achieves an inference throughput of 7 fps on a single A100 GPU.

\textbf{Observations And Remarks.} 
Employing separate backbones for each input branch enables modality-specific feature learning. However, this approach does not inherently outperform a single-backbone architecture with concatenated inputs. Moreover, achieving optimal performance likely requires extensive hyperparameter tuning and a refined training strategy. The exclusive use of the MultiAqua dataset may further limit the model's ability to fully exploit the potential of multi-modal fusion, indicating that a larger and more diverse dataset may be required. Finally, the inherently low contrast of thermal imagery presents an additional challenge, suggesting that more sophisticated preprocessing techniques warrant further investigation.

\newpage

{\small
\bibliographystyle{ieee_fullname}
\bibliography{egbib}

@STRING{IROS = {Int. Conf. Intell. Robots and Systems}}

@article{KristanPAMI2016,
  title       = {A Novel Performance Evaluation Methodology for Single-Target Trackers},
  author      = {Matej Kristan and Jiri Matas and Ale\v{s} Leonardis and Tomas Voj{\'{i}\~{r}} and Roman Pflugfelder and Gustavo Fern{\'{a}}ndez and Georg Nebehay and Fatih Porikli and Luka \v{C}ehovin},
  journal     = {IEEE Transactions on Pattern Analysis and Machine Intelligence},
  volume      = {38},
  number      = {11},
  pages       = {2137--2155},
  year        = {2016},
  publisher   = {IEEE}
}

@inproceedings{varga2022seadronessee,
  title={Seadronessee: A maritime benchmark for detecting humans in open water},
  author={Varga, Leon Amadeus and Kiefer, Benjamin and Messmer, Martin and Zell, Andreas},
  booktitle={Proceedings of the IEEE/CVF Winter Conference on Applications of Computer Vision},
  pages={2260--2270},
  year={2022}
}

@article{prasad2019object,
  title={Are object detection assessment criteria ready for maritime computer vision?},
  author={Prasad, Dilip K and Dong, Huixu and Rajan, Deepu and Quek, Chai},
  journal={IEEE Transactions on Intelligent Transportation Systems},
  volume={21},
  number={12},
  pages={5295--5304},
  year={2019},
  publisher={IEEE}
}

@article{kanjir2018vessel,
  title={Vessel detection and classification from spaceborne optical images: A literature survey},
  author={Kanjir, Ur{\v{s}}ka and Greidanus, Harm and O{\v{s}}tir, Kri{\v{s}}tof},
  journal={Remote sensing of environment},
  volume={207},
  pages={1--26},
  year={2018},
  publisher={Elsevier}
}

@article{gallego2019detection,
  title={Detection of bodies in maritime rescue operations using unmanned aerial vehicles with multispectral cameras},
  author={Gallego, Antonio-Javier and Pertusa, Antonio and Gil, Pablo and Fisher, Robert B},
  journal={Journal of Field Robotics},
  volume={36},
  number={4},
  pages={782--796},
  year={2019},
  publisher={Wiley Online Library}
}

@misc{luxonis_oak4d,
  author       = {{Luxonis}},
  title        = {{OAK4-D}},
  howpublished = {\url{https://docs.luxonis.com/hardware/products/OAK\%204\%20D}},
  note         = {Luxonis documentation. Accessed: 2026-04-08}
}

@article{kiefer2021leveraging,
  title={Leveraging Synthetic Data in Object Detection on Unmanned Aerial Vehicles},
  author={Kiefer, Benjamin and Ott, David and Zell, Andreas},
  journal={arXiv preprint arXiv:2112.12252},
  year={2021}
}

@InProceedings{Kiefer_2026_CVPR,
    author    = {Benjamin Kiefer and Dominik Hildebrand and Rafia Rahim and Mahmut Karaaslan and Michael DeFilippo and Ersin Kaya and Andreas Zell},
    title     = {Real-Time Radar--Vision Association via Monocular Distance Estimation},
    booktitle = {Proceedings of the IEEE/CVF Conference on Computer Vision and Pattern Recognition (CVPR) Workshops},
    month     = {June},
    year      = {2026}
}

@InProceedings{Kiefer_2025_WACV,
    author    = {Kiefer, Benjamin and Zust, Lojze and Kristan, Matej and Pers, Janez and Tersek, Matija and Mudenagudi, Uma and Desai, Chaitra and Wiliem, Arnold and Kreis, Marten and Akalwadi, Nikhil and Zhong, Zhiqiang and Zhang, Zhe and Liu, Sujie and Chen, Xuran and Yang, Yang and Fabijanic, Matej and Ferreira, Fausto and Lee, Seongju and Yao, Shanliang and Kumar, Himanshu and Marcus, Aurelius and Novak, Gregor and Feng, Yuan and Cheng, Annie and Nguyen, Thien and Sheikh, Jannik and Saric, Josip and Li, Zhuoxiao and Lu, Yutang and Lin, Yipeng and Yang, Xiang and Cheng, Ching-Heng and Awad, Ali and Muhovi\v{c}, Jon and Quan, Yitong and Lee, Junseok and Lee, Kyoobin and Guan, Runwei and Huang, Xiaoyu and Ni, Yi and Lin, Tzu-Yu and Lee, Chia-Ming and Hsu, Chih-Chung and Michel, Andreas and Gross, Wolfgang and Jiang, Nan and Feng, Fei and Lucas, Evan and Saleem, Ashraf and Lin, Yu-Fan and Weinmann, Martin},
    title     = {3rd Workshop on Maritime Computer Vision (MaCVi) 2025: Challenge Results},
    booktitle = {Proceedings of the Winter Conference on Applications of Computer Vision (WACV) Workshops},
    month     = {February},
    year      = {2025},
    pages     = {1542-1569}
}

@article{bloisi2014background,
  title={Background modeling and foreground detection for maritime video surveillance},
  author={Bloisi, Domenico},
  journal={Chapter in Handbook on Background Modeling and Foreground Detection for Video Surveillance: Traditional and Recent Approaches, Implementations, Benchmarking and Evaluation, Chapman and Hall/CRC},
  pages={14--1},
  year={2014}
}

@ARTICLE{Bovcon2021,
  author={Bovcon, Borja and Kristan, Matej},
  journal={IEEE Transactions on Cybernetics}, 
  title={{WaSR--A Water Segmentation and Refinement Maritime Obstacle Detection Network}}, 
  year={2021},
  volume={},
  number={},
  pages={1-14},
  doi={10.1109/TCYB.2021.3085856}}

@InProceedings{Zust2022Learning,
  title={Learning Maritime Obstacle Detection from Weak Annotations by Scaffolding},
  author={{\v{Z}}ust, Lojze and Kristan, Matej},
  booktitle={Proceedings of the IEEE/CVF Winter Conference on Applications of Computer Vision},
  pages={955--964},
  year={2022}
}

@inproceedings{bovcon2019mastr,
  title={The MaSTr1325 Dataset for Training Deep USV Obstacle Detection Models},
  author={Bovcon, Borja and Muhovi{\v{c}}, Jon and Per{\v{s}}, Janez and Kristan, Matej},
  booktitle=IROS,
  pages={3431--3438},
  year={2019},
  organization={IEEE}
}

@misc{lookout,
  title = {{LOOKOUT AI System}},
  howpublished ={\url{https://www.getalookout.com/}},
  note = {Accessed: 2023-11-18}
}

@inproceedings{deng2009imagenet,
  title={Imagenet: A large-scale hierarchical image database},
  author={Deng, Jia and Dong, Wei and Socher, Richard and Li, Li-Jia and Li, Kai and Fei-Fei, Li},
  booktitle={2009 IEEE conference on computer vision and pattern recognition},
  pages={248--255},
  year={2009},
  organization={Ieee}
}

@inproceedings{loshchilov2018decoupled,
title={Decoupled Weight Decay Regularization},
author={Ilya Loshchilov and Frank Hutter},
booktitle={International Conference on Learning Representations},
year={2019},
url={https://openreview.net/forum?id=Bkg6RiCqY7},
}

@inproceedings{gupta2019lvis,
  title={Lvis: A dataset for large vocabulary instance segmentation},
  author={Gupta, Agrim and Dollar, Piotr and Girshick, Ross},
  booktitle={Proceedings of the IEEE Conference on Computer Vision and Pattern Recognition},
  pages={5356--5364},
  year={2019}
}

@InProceedings{Kiefer_2023_WACV,
    author    = {Kiefer, Benjamin and Kristan, Matej and Per\v{s}, Janez and \v{Z}ust, Lojze and Poiesi, Fabio and Andrade, Fabio and Bernardino, Alexandre and Dawkins, Matthew and Raitoharju, Jenni and Quan, Yitong and Atmaca, Adem and H\"ofer, Timon and Zhang, Qiming and Xu, Yufei and Zhang, Jing and Tao, Dacheng and Sommer, Lars and Spraul, Raphael and Zhao, Hangyue and Zhang, Hongpu and Zhao, Yanyun and Augustin, Jan Lukas and Jeon, Eui-ik and Lee, Impyeong and Zedda, Luca and Loddo, Andrea and Di Ruberto, Cecilia and Verma, Sagar and Gupta, Siddharth and Muralidhara, Shishir and Hegde, Niharika and Xing, Daitao and Evangeliou, Nikolaos and Tzes, Anthony and Bartl, Vojt\v{e}ch and \v{S}pa\v{n}hel, Jakub and Herout, Adam and Bhowmik, Neelanjan and Breckon, Toby P. and Kundargi, Shivanand and Anvekar, Tejas and Tabib, Ramesh Ashok and Mudenagudi, Uma and Vats, Arpita and Song, Yang and Liu, Delong and Li, Yonglin and Li, Shuman and Tan, Chenhao and Lan, Long and Somers, Vladimir and De Vleeschouwer, Christophe and Alahi, Alexandre and Huang, Hsiang-Wei and Yang, Cheng-Yen and Hwang, Jenq-Neng and Kim, Pyong-Kun and Kim, Kwangju and Lee, Kyoungoh and Jiang, Shuai and Li, Haiwen and Ziqiang, Zheng and Vu, Tuan-Anh and Nguyen-Truong, Hai and Yeung, Sai-Kit and Jia, Zhuang and Yang, Sophia and Hsu, Chih-Chung and Hou, Xiu-Yu and Jhang, Yu-An and Yang, Simon and Yang, Mau-Tsuen},
    title     = {1st Workshop on Maritime Computer Vision (MaCVi) 2023: Challenge Results},
    booktitle = {Proceedings of the IEEE/CVF Winter Conference on Applications of Computer Vision (WACV) Workshops},
    month     = {January},
    year      = {2023},
    pages     = {265-302}
}

@InProceedings{Kiefer_2024_WACV,
    author    = {Kiefer, Benjamin and \v{Z}ust, Lojze and Kristan, Matej and Per\v{s}, Janez and Ter\v{s}ek, Matija and Wiliem, Arnold and Messmer, Martin and Yang, Cheng-Yen and Huang, Hsiang-Wei and Jiang, Zhongyu and Kuo, Heng-Cheng and Mei, Jie and Hwang, Jenq-Neng and Stadler, Daniel and Sommer, Lars and Huang, Kaer and Zheng, Aiguo and Chong, Weitu and Lertniphonphan, Kanokphan and Xie, Jun and Chen, Feng and Li, Jian and Wang, Zhepeng and Zedda, Luca and Loddo, Andrea and Di Ruberto, Cecilia and Vu, Tuan-Anh and Nguyen-Truong, Hai and Ha, Tan-Sang and Pham, Quan-Dung and Yeung, Sai-Kit and Feng, Yuan and Thien, Nguyen Thanh and Tian, Lixin and Michel, Andreas and Gross, Wolfgang and Weinmann, Martin and Carrillo-Perez, Borja and Klein, Alexander and Alex, Antje and Solano-Carrillo, Edgardo and Steiniger, Yannik and Rodriguez, Angel Bueno and Kuan, Sheng-Yao and Ho, Yuan-Hao and Sattler, Felix and Fabijani\'c, Matej and \v{S}imunec, Magdalena and Kapetanovi\'c, Nadir},
    title     = {2nd Workshop on Maritime Computer Vision (MaCVi) 2024: Challenge Results},
    booktitle = {Proceedings of the IEEE/CVF Winter Conference on Applications of Computer Vision (WACV) Workshops},
    month     = {January},
    year      = {2024},
    pages     = {869-891}
}

@misc{kiefer2025approximatesupervisedobjectdistance,
      title={Approximate Supervised Object Distance Estimation on Unmanned Surface Vehicles}, 
      author={Benjamin Kiefer and Yitong Quan and Andreas Zell},
      year={2025},
      eprint={2501.05567},
      archivePrefix={arXiv},
      primaryClass={cs.CV},
      url={https://arxiv.org/abs/2501.05567}, 
}

@InProceedings{Zust2023LaRS,
  title={{LaRS}: A Diverse Panoptic Maritime Obstacle Detection Dataset and Benchmark},
  author={{\v{Z}}ust, Lojze and Per{\v{s}}, Janez and Kristan, Matej},
  booktitle={International Conference on Computer Vision (ICCV)},
  year={2023}
}

@article{Zust2024PanSR,
  title={{PanSR}: An Object-Centric Mask Transformer for Panoptic Segmentation},
  author={{\v{Z}}ust, Lojze and Kristan, Matej},
  journal={arXiv preprint arXiv:2412.10589},
  year={2024}
}

@inproceedings{cheng2021mask2former,
  title={Masked-attention Mask Transformer for Universal Image Segmentation},
  author={Bowen Cheng and Ishan Misra and Alexander G. Schwing and Alexander Kirillov and Rohit Girdhar},
  booktitle={IEEE Conference on Computer Vision and Pattern Recognition (CVPR)},
  year={2022}
}

@misc{mmseg2020,
    title={{MMSegmentation}: OpenMMLab Semantic Segmentation Toolbox and Benchmark},
    author={MMSegmentation Contributors},
    howpublished = {\url{https://github.com/open-mmlab/mmsegmentation}},
    year={2020}
}

@inproceedings{liu2021swin,
  title={Swin transformer: Hierarchical vision transformer using shifted windows},
  author={Liu, Ze and Lin, Yutong and Cao, Yue and Hu, Han and Wei, Yixuan and Zhang, Zheng and Lin, Stephen and Guo, Baining},
  booktitle={Proceedings of the IEEE/CVF International Conference on Computer Vision},
  pages={10012--10022},
  year={2021}
}

@misc{Li2022Mask,
  title = {Mask {{DINO}}: {{Towards A Unified Transformer-based Framework}} for {{Object Detection}} and {{Segmentation}}},
  shorttitle = {Mask {{DINO}}},
  author = {Li, Feng and Zhang, Hao and {xu}, Huaizhe and Liu, Shilong and Zhang, Lei and Ni, Lionel M. and Shum, Heung-Yeung},
  year = {2022},
  month = dec,
  number = {arXiv:2206.02777},
  eprint = {2206.02777},
  primaryclass = {cs},
  publisher = {arXiv},
  urldate = {2023-05-09},
  archiveprefix = {arXiv},
  langid = {english}
}

@article{oquabdinov2,
  title={DINOv2: Learning Robust Visual Features without Supervision},
  author={Oquab, Maxime and Darcet, Timoth{\'e}e and Moutakanni, Th{\'e}o and Vo, Huy V and Szafraniec, Marc and Khalidov, Vasil and Fernandez, Pierre and HAZIZA, Daniel and Massa, Francisco and El-Nouby, Alaaeldin and others},
  journal={Transactions on Machine Learning Research},
  year={2023}
}

@article{yao2024waterscenes,
  title={Waterscenes: A multi-task 4d radar-camera fusion dataset and benchmarks for autonomous driving on water surfaces},
  author={Yao, Shanliang and Guan, Runwei and Wu, Zhaodong and Ni, Yi and Huang, Zile and Liu, Ryan Wen and Yue, Yong and Ding, Weiping and Lim, Eng Gee and Seo, Hyungjoon and others},
  journal={IEEE Transactions on Intelligent Transportation Systems},
  year={2024},
  publisher={IEEE}
}

@article{rsuigm,
author = {Desai, Chaitra and Benur, Sujay and Patil, Ujwala and Mudenagudi, Uma},
title = {RSUIGM: Realistic Synthetic Underwater Image Generation with Image Formation Model},
year = {2024},
issue_date = {January 2025},
publisher = {Association for Computing Machinery},
address = {New York, NY, USA},
volume = {21},
number = {1},
issn = {1551-6857},
url = {https://doi.org/10.1145/3656473},
doi = {10.1145/3656473},
journal = {ACM Trans. Multimedia Comput. Commun. Appl.},
month = dec,
articleno = {10},
numpages = {22}
}

@inproceedings{Kirillov2019Panoptic,
  title = {Panoptic Segmentation},
  booktitle = {Proceedings of the {{IEEE Computer Society Conference}} on {{Computer Vision}} and {{Pattern Recognition}}},
  author = {Kirillov, Alexander and He, Kaiming and Girshick, Ross and Rother, Carsten and Dollar, Piotr},
  year = {2019},
  month = jun,
  volume = {2019-June},
  eprint = {1801.00868},
  pages = {9396--9405},
  publisher = {IEEE Computer Society},
  urldate = {2021-01-13},
  archiveprefix = {arXiv}
}

@misc{Jocher_Ultralytics_YOLO_2023,
author = {Jocher, Glenn and Chaurasia, Ayush and Qiu, Jing},
license = {AGPL-3.0},
month = jan,
title = {{Ultralytics YOLO}},
url = {https://github.com/ultralytics/ultralytics},
version = {8.0.0},
year = {2023}
}

@inproceedings{gorczak2025maritime,
  title={Maritime Collision Avoidance Dataset Germany, English Channel, and The Netherlands},
  author={Gorczak, Philipp and L{\"u}bcke, Thomas and Portier, Martin and Schmid, Helmut},
  booktitle={Journal of Physics: Conference Series},
  volume={3123},
  pages={012024},
  year={2025},
  organization={IOP Publishing}
}

@article{solovyev2021wbf,
  title={Weighted boxes fusion: Ensembling boxes from different object detection models},
  author={Solovyev, Roman and Wang, Weimin and Gabruseva, Tatiana},
  journal={Image and Vision Computing},
  volume={107},
  pages={104117},
  year={2021},
  publisher={Elsevier}
}

@article{chen2023mixpl,
  title={Mixed pseudo labels for semi-supervised object detection},
  author={Chen, Zeming and Zhang, Wenwei and Wang, Xinjiang and Chen, Kai and Wang, Zhi},
  journal={arXiv preprint arXiv:2312.07006},
  year={2023}
}

@inproceedings{zong2023codino,
  title={DETRs with collaborative hybrid assignments training},
  author={Zong, Zhuofan and Song, Guanglu and Liu, Yu},
  booktitle={ICCV},
  year={2023}
}

@article{lyu2022rtmdet,
  title={RTMDet: An empirical study of designing real-time object detectors},
  author={Lyu, Chengqi and Zhang, Wenwei and Huang, Haian and Zhou, Yue and Wang, Yudong and Liu, Yanyi and Zhang, Shilong and Chen, Kai},
  journal={arXiv preprint arXiv:2212.07784},
  year={2022}
}

@inproceedings{robinson2026rfdetr,
  title={RF-DETR: Neural architecture search for real-time detection transformers},
  author={Robinson, Isaac and Robicheaux, Peter and Popov, Matvei and Ramanan, Deva and Peri, Neehar},
  booktitle={ICLR},
  year={2026}
}

@inproceedings{zhang2023dino,
  title={DINO: DETR with improved denoising anchor boxes for end-to-end object detection},
  author={Zhang, Hao and Li, Feng and Liu, Shilong and Zhang, Lei and Su, Hang and Zhu, Jun and Ni, Lionel M and Shum, Heung-Yeung},
  booktitle={ICLR},
  year={2023}
}

@inproceedings{zhang2023ddqdetr,
  title={Dense distinct query for end-to-end object detection},
  author={Zhang, Shilong and Wang, Xinjiang and Wang, Jiaqi and Pang, Jiangmiao and Lyu, Chengqi and Zhang, Wenwei and Luo, Ping and Chen, Kai},
  booktitle={CVPR},
  year={2023}
}

@article{simeoni2025dinov3,
  title={Dinov3},
  author={Sim{\'e}oni, Oriane and Vo, Huy V and Seitzer, Maximilian and Baldassarre, Federico and Oquab, Maxime and Jose, Cijo and Khalidov, Vasil and Szafraniec, Marc and Yi, Seungeun and Ramamonjisoa, Micha{\"e}l and others},
  journal={arXiv preprint arXiv:2508.10104},
  year={2025}
}

@inproceedings{liu2022convnet,
  title={A ConvNet for the 2020s},
  author={Liu, Zhuang and Mao, Hanzi and Wu, Chao-Yuan and Feichtenhofer, Christoph and Darrell, Trevor and Xie, Saining},
  booktitle={Proceedings of the IEEE/CVF Conference on Computer Vision and Pattern Recognition (CVPR)},
  pages={11976--11986},
  year={2022}
}

@inproceedings{li2022exploring,
  title={Exploring plain vision transformer backbones for object detection},
  author={Li, Yanghao and Mao, Hanzi and Girshick, Ross and He, Kaiming},
  booktitle={European conference on computer vision},
  pages={280--296},
  year={2022},
  organization={Springer}
}

@inproceedings{kerssies2025your,
  title={Your vit is secretly an image segmentation model},
  author={Kerssies, Tommie and Cavagnero, Niccolo and Hermans, Alexander and Norouzi, Narges and Averta, Giuseppe and Leibe, Bastian and Dubbelman, Gijs and De Geus, Daan},
  booktitle={Proceedings of the computer vision and pattern recognition conference},
  pages={25303--25313},
  year={2025}
}

@article{huang2025real,
  title={Real-time object detection meets DINOv3},
  author={Huang, Shihua and Hou, Yongjie and Liu, Longfei and Yu, Xuanlong and Shen, Xi},
  journal={arXiv preprint arXiv:2509.20787},
  year={2025}
}

@article{ridnik2021imagenet,
  title={Imagenet-21k pretraining for the masses},
  author={Ridnik, Tal and Ben-Baruch, Emanuel and Noy, Asaf and Zelnik-Manor, Lihi},
  journal={arXiv preprint arXiv:2104.10972},
  year={2021}
}

@ARTICLE{wang2026rsosnet,
  author={Wang, Ning and Feng, Yuan and Tian, Lixin and Wei, Yi},
  journal={IEEE Transactions on Intelligent Transportation Systems}, 
  title={RSOS-Net: Real-Time Surface Obstacle Segmentation Network for Uncrewed Waterborne Vehicles}, 
  year={2026},
  volume={27},
  number={1},
  pages={1052-1065},
  doi={10.1109/TITS.2025.3628677}}

@article{yolox2021,
  title={YOLOX: Exceeding YOLO Series in 2021},
  author={Ge, Zheng and Liu, Songtao and Wang, Feng and Li, Zeming and Sun, Jian},
  journal={arXiv preprint arXiv:2107.08430},
  year={2021}
}

@inproceedings{xu2023pidnet,
  title={{PIDNet}: A Real-Time Semantic Segmentation Network Inspired by {PID} Controllers},
  author={Xu, Jiacong and Xiong, Zixiang and Bhattacharyya, Shankar P.},
  booktitle={Proceedings of the IEEE/CVF Conference on Computer Vision and Pattern Recognition (CVPR)},
  pages={19529--19539},
  year={2023}
}

@inproceedings{ghiasi2021copypaste,
  title={Simple Copy-Paste is a Strong Data Augmentation Method for Instance Segmentation},
  author={Ghiasi, Golnaz and Cui, Yin and Srinivas, Aravind and Qian, Rui and Lin, Tsung-Yi and Cubuk, Ekin D. and Le, Quoc V. and Zoph, Barret},
  booktitle={Proceedings of the IEEE/CVF Conference on Computer Vision and Pattern Recognition (CVPR)},
  pages={2918--2928},
  year={2021}
}

@article{lambert2022rosebud,
  title={{ROSEBUD}: A Deep Fluvial Segmentation Dataset for Monocular Vision-Based River Navigation and Obstacle Avoidance},
  author={Lambert, Reeve and Chavez-Galaviz, Jalil and Li, Jianwen and Mahmoudian, Nina},
  journal={Sensors},
  volume={22},
  number={13},
  pages={4681},
  year={2022},
  publisher={MDPI},
  doi={10.3390/s22134681}
}

@misc{iancu2024multiaqua,
  title={{MULTIAQUA}: A Multimodal Maritime Dataset and Robust Training Strategies for Multimodal Semantic Segmentation},
  author={Muhovi{\v{c}}, Jon and Per{\v{s}}, Janez},
  year={2025},
  eprint={2512.17450},
  archivePrefix={arXiv},
  primaryClass={cs.CV}
}

@misc{kreis2025realtimefusionvisualchart,
      title={Real-Time Fusion of Visual and Chart Data for Enhanced Maritime Vision}, 
      author={Marten Kreis and Benjamin Kiefer},
      year={2025},
      eprint={2507.13880},
      archivePrefix={arXiv},
      primaryClass={cs.CV},
      url={https://arxiv.org/abs/2507.13880}, 
}

@inproceedings{michel2023dustnet,
  title={Dustnet: Attention to dust},
  author={Michel, Andreas and Weinmann, Martin and Schenkel, Fabian and Gomez, Tomas and Falvey, Mark and Schmitz, Rainer and Middelmann, Wolfgang and Hinz, Stefan},
  booktitle={DAGM German Conference on Pattern Recognition},
  pages={211--226},
  year={2023},
  organization={Springer}
}

@article{michel2025dustnet++,
  title={Dustnet++: Deep learning-based visual regression for dust density estimation},
  author={Michel, Andreas and Weinmann, Martin and Kuester, Jannick and Alnasser, Faisal and Gomez, Tomas and Falvey, Mark and Schmitz, Rainer and Middelmann, Wolfgang and Hinz, Stefan},
  journal={International Journal of Computer Vision},
  volume={133},
  number={7},
  pages={4220--4244},
  year={2025},
  publisher={Springer}
}

@article{loshchilov2017decoupled,
  title={Decoupled weight decay regularization},
  author={Loshchilov, Ilya and Hutter, Frank},
  journal={arXiv preprint arXiv:1711.05101},
  year={2017}
}

@incollection{zuiderveld1994contrast,
  title={Contrast limited adaptive histogram equalization},
  author={Zuiderveld, Karel},
  booktitle={Graphics gems IV},
  pages={474--485},
  year={1994}
}

@article{liao2025memorysam,
  author       = {C. Liao and X. Zheng and Y. Lyu and H. Xue and Y. Cao and J. Wang and K. Yang and X. Hu},
  title        = {MemorySAM: Memorize Modalities and Semantics with Segment Anything Model 2 for Multi-Modal Semantic Segmentation},
  journal      = {arXiv preprint arXiv:2503.06700},
  year         = {2025}
}

@article{ravi2024sam2,
  author       = {N. Ravi and V. Gabber and Y.-T. Hu and R. Hu and C. Ryali and T. Ma and H. Khedr and R. R{\"a}dle and C. Rolber and L. Gustafson and E. Mintun and J. Pan and K. V. Alwala and N. Carion and C.-Y. Wu and R. Girshick and P. Doll{\'a}r and C. Feichtenhofer},
  title        = {SAM 2: Segment Anything in Images and Videos},
  journal      = {arXiv preprint arXiv:2408.00714},
  year         = {2024}
}

@inproceedings{hu2022lora,
  author       = {E. J. Hu and Y. Shen and P. Wallis and Z. Allen-Zhu and Y. Li and S. Wang and L. Wang and W. Chen},
  title        = {LoRA: Low-Rank Adaptation of Large Language Models},
  booktitle    = {International Conference on Learning Representations (ICLR)},
  year         = {2022}
}

@inproceedings{puigcerver2024softmoe,
  author       = {J. Puigcerver and C. Riquelme and B. Mustafa and N. Houlsby},
  title        = {From Sparse to Soft Mixtures of Experts},
  booktitle    = {International Conference on Learning Representations (ICLR)},
  year         = {2024}
}

@inproceedings{xu2025physaug,
  author       = {X. Xu and J. Yang and W. Shi and S. Ding and L. Luo and J. Liu},
  title        = {PhysAug: A Physical-Guided and Frequency-Based Data Augmentation for Single-Domain Generalized Object Detection},
  booktitle    = {Proceedings of the AAAI Conference on Artificial Intelligence},
  volume       = {39},
  pages        = {21815--21823},
  year         = {2025}
}
}

\end{document}